\documentclass[lettersize,journal]{IEEEtran}
\usepackage{amsmath,amsfonts}
\usepackage{algorithmic}
\usepackage{algorithm}
\usepackage{array}
\usepackage[caption=false,font=normalsize,labelfont=sf,textfont=sf]{subfig}
\usepackage{textcomp}
\usepackage{stfloats}
\usepackage{url}
\usepackage{verbatim}
\usepackage{graphicx}
\usepackage{cite}
\usepackage{hyperref}
\usepackage{mciteplus}
\usepackage{caption}
\usepackage{booktabs}       % professional-quality tables
\usepackage{amssymb}
\usepackage{color}
\usepackage{multirow}

\usepackage{cuted}

\def\Secref#1{Section~\ref{#1}}
\def\eqref#1{equation~\ref{#1}}
\newtheorem{proposition}{Proposition}

\begin{document}

\title{MInCo: Mitigating Information Conflicts in Distracted Visual Model-based Reinforcement Learning}

\author{Shiguang Sun, Hanbo Zhang, ~\IEEEmembership{Member,~IEEE,} Zeyang Liu, Xinrui Yang, Lipeng Wan, Xingyu Chen, ~\IEEEmembership{Member,~IEEE,} Xuguang Lan, ~\IEEEmembership{Senior Member,~IEEE,}
% IEEE Publication Technology,~\IEEEmembership{Staff,~IEEE,}
       % <-this stops a space
\thanks{This work was supported in part by NSFC under grant No.62125305, No. U23A20339, No.62088102, No. 62203348. (Corresponding authors: Xingyu Chen; Xuguang Lan.)}% <-this % stops a space
\thanks{Shiguang Sun, Zeyang Liu, Xinrui Yang, Lipeng Wan, Xingyu Chen, and Xuguang Lan are with the National Key Laboratory of Human-Machine Hybrid Augmented Intelligence, National Engineering Research Center for Visual Information and Applications, and Institute of Artificial Intelligence and Robotics, Xi’an Jiaotong University, No.28 West Xianning Road, Xi’an, 710049, P. R. China (e-mail: ssg2019@stu.xjtu.edu.cn, zeyang.liu@stu.xjtu.edu.cn, xinrui.yang@stu.xjtu.edu.cn, wanlipeng@stu.xjtu.edu.cn, chenxingyu\_1990@xjtu.edu.cn, xglan@mail.xjtu.edu.cn).}% <-this % stops a space
\thanks{Hanbo Zhang is with the National University of Singapore, 21 Lower Kent Ridge Road Singapore, 119077, Singapore (e-mail: zhanghb@comp.nus.edu.sg)}

% \thanks{Manuscript received }
}

% \markboth{IEEE TRANSACTIONS ON CYBERNETICS}%


\maketitle
\begin{abstract}
 Existing visual model-based reinforcement learning (MBRL) algorithms with observation reconstruction often suffer from information conflicts, making it difficult to learn compact representations and hence result in less robust policies, especially in the presence of task-irrelevant visual distractions.
    In this paper, we first reveal that the information conflicts in current visual MBRL algorithms stem from visual representation learning and latent dynamics modeling with an information-theoretic perspective.
    Based on this finding, we present a new algorithm to resolve information conflicts for visual MBRL, named MInCo,
    which mitigates information conflicts by leveraging negative-free contrastive learning, aiding in learning invariant representation and robust policies despite noisy observations. 
    To prevent the dominance of visual representation learning, we introduce time-varying reweighting to bias the learning towards dynamics modeling as training proceeds.
    We evaluate our method on several robotic control tasks with dynamic background distractions.
    Our experiments demonstrate that MInCo learns invariant representations against background noise and consistently outperforms current state-of-the-art visual MBRL methods. Code is available at \url{https://github.com/ShiguangSun/minco}.
\end{abstract}


\section{Introduction}\label{intro}
\IEEEPARstart{A}{mong} various forms of machine learning, reinforcement learning (RL), which models decision-making through trial and error, is one of the approaches most similar to human learning \cite{sutton2018reinforcement}. 
Deep reinforcement learning (DRL) utilizes deep neural networks to tackle high-dimensional tasks, achieving notable success in fields such as robotic manipulation \cite{8890006,10684782,9366328}, autonomous driving \cite{9582785,9478933,9537641}, video games \cite{DBLP:journals/nature/MnihKSRVBGRFOPB15}, Go \cite{DBLP:journals/nature/SilverSSAHGHBLB17}, and others \cite{9596578}.
% \IEEEPARstart{R}{ecently}, deep reinforcement learning (DRL) has advanced rapidly, achieving impressive results in various complex tasks such as robotic manipulation \cite{DBLP:journals/tcyb/XiangS21, DBLP:journals/tcyb/WangZLKS24}, autonomous driving \cite{DBLP:journals/tcyb/ZhangLLGGCL24}, video games \cite{DBLP:journals/nature/MnihKSRVBGRFOPB15, DBLP:journals/tcyb/QuOG22}, Go \cite{DBLP:journals/nature/SilverSSAHGHBLB17}, and others \cite{DBLP:journals/tcyb/HouCLHGKJ24}. Among various forms of machine learning, reinforcement learning (RL) is one of the approaches most similar to human learning \cite{sutton2018reinforcement}.
% Planning actions based on predictions of environments is one of the fundamental abilities of human-level intelligence, particularly useful in long-horizon tasks. 
Another fundamental human ability is planning actions based on predictions of the environment, which is especially useful in long-term tasks.
% To achieve this, model-based reinforcement learning (MBRL) \cite{sutton2018reinforcement}
Model-based reinforcement learning (MBRL), designed to mimic this ability, learns a dynamics model to predict the future and then performs policy optimization \cite{DBLP:conf/iclr/HafnerLB020,DBLP:conf/iclr/HafnerL0B21,DBLP:journals/corr/abs-2301-04104} or planning \cite{DBLP:journals/nature/SchrittwieserAH20,DBLP:conf/nips/YeLKAG21,DBLP:conf/icml/HansenSW22,DBLP:journals/corr/abs-2310-16828} to determine the optimal action sequence.
However, modeling environmental dynamics from high-dimensional visual observations remains challenging, particularly when faced with noise and distractions in the background.

\begin{figure*}
	\centering
	\includegraphics[width=\textwidth]{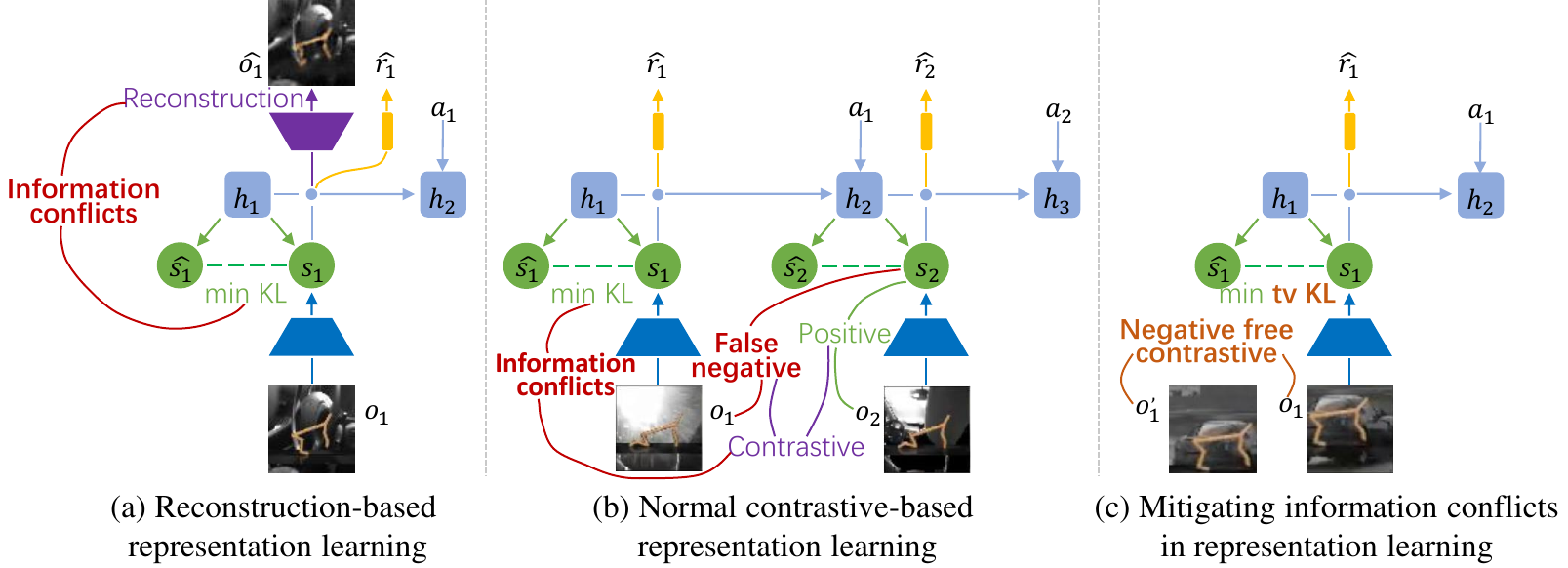}
	\caption{Representation learning designs. Here, \(o\) is observation, \( \hat{s} \) is the prior latent state, \(s\) is the posterior latent state conditioned on the observation, \(h\) represents the intermediate output of a neural network (e.g., the hidden state in a GRU), \( \hat{o} \) and \(\hat{r}\) are the reconstructed reward and observation, respectively, and \(a\) is action in training data. (a) In reconstruction-based methods, image reconstruction and minimizing KL divergence cause information conflicts. (b) In contrastive-based methods, the InfoNCE \cite{DBLP:journals/corr/abs-1807-03748} contrastive loss also introduces information conflicts and causes false negatives. (c) Our approach uses negative-free contrastive loss and time-varying KL divergence, mitigating information conflicts and avoiding false negatives.}
    \label{fig:method compare}
    \vspace{-10pt}
\end{figure*}

It has been shown that learning robust representations is crucial when dealing with such high-dimensional and noisy inputs \cite{DBLP:conf/iclr/YamadaPGL22,DBLP:conf/ijcai/KimHK22,DBLP:conf/aaai/0008ZY023a}.
Recent works have widely explored the learning of representations for visual observations of environmental dynamics \cite{DBLP:conf/iclr/HafnerLB020,DBLP:conf/iclr/HafnerL0B21,DBLP:conf/corl/0006CHL20,DBLP:conf/acml/WangYWL22,DBLP:conf/icra/OkadaT21,DBLP:conf/icml/DengJA22,DBLP:conf/iros/OkadaT22,DBLP:conf/nips/Pan0WY22,DBLP:conf/icml/FuYAJ21,DBLP:conf/icml/0001D0IZT22}.
They can be mainly categorized into two classes: reconstruction-based approaches \cite{DBLP:conf/icml/HafnerLFVHLD19,DBLP:conf/iclr/HafnerLB020,DBLP:conf/iclr/HafnerL0B21,DBLP:journals/corr/abs-2301-04104,DBLP:conf/nips/Pan0WY22,DBLP:conf/icml/0001D0IZT22}, as shown in Fig. \ref{fig:method compare} (a), and contrastive-based approaches \cite{DBLP:conf/corl/0006CHL20,DBLP:conf/acml/WangYWL22,DBLP:conf/icra/OkadaT21,DBLP:conf/iros/OkadaT22}, as shown in Fig. \ref{fig:method compare} (b).
% Dreamer learns representations by minimizing reconstruction error. However, we find that from an information-theoretic perspective, there is an information conflict between Dreamer's reconstruction objective and the latent dynamics objective. 
Nevertheless, in this paper, we reveal the fact that most of them suffer from \textit{information conflicts}.
% stemming from visual representation learning and dynamics modeling.
We review and re-formalize the problem of learning low-dimensional state representations and latent dynamics models in visual MBRL from an information-theoretic perspective.
It highlights that the information conflicts in existing visual MBRL algorithms stem from the visual reconstruction loss and latent dynamics loss in the model's training objectives.
% It points out that information conflicts of existing visual MBRL algorithms lie in the visual reconstruction loss and latent dynamics loss of the model's training objectives.
Such information conflicts can harm the learning of compact and robust representations, especially with distracted or noisy observations.

Therefore, designing a reconstruction-free MBRL algorithm that can avoid information conflicts and achieve robust representation learning is a critical challenge.
To achieve this, we propose a novel algorithm named MInCo, aiming at \textbf{M}itigating the \textbf{In}formation \textbf{Co}nflicting problem and learning robust representations for environmental dynamics.
As shown in Fig. \ref{fig:method compare} (c), MInCo mitigates information conflicts by leveraging negative-free contrastive learning without image reconstruction.
However, simply replacing the reconstruction loss with a negative-free contrastive loss alters the relative contributions of different parts of the model's loss function, which can negatively impact representation learning.
% To prevent visual representation learning from dominating the entire training progress,
To solve this challenge, we further introduce a time-varying reweighting strategy, which biases the learning towards dynamics modeling gradually as training proceeds.
As a result, MInCo learns robust, compact, and generalizable representations for environmental dynamics despite noisy observations and distractions in the backgrounds.

Specifically, the main contributions of this work are summarized as follows:
\begin{itemize}
    \vspace{-5pt}
    \item  We reinterpret visual model-based reinforcement learning from an information-theoretic perspective, pointing out the information conflict problem therein. 
% Additionally, we discover the problem of positive-negative sample confusion when applying contrastive learning loss InfoNCE \cite{DBLP:journals/corr/abs-1807-03748} in reinforcement learning.
    \item We propose a novel representation learning algorithm, MInCo, which utilizes a time-varying dynamics loss and SimSiam contrastive loss to mitigate information conflict. Furthermore, replacing the reconstruction loss with the SimSiam loss helps avoid positive-negative sample confusion caused by the InfoNCE loss.
    \item  We evaluate MinCo on continuous control tasks with complex background distractions, and the results demonstrate that our method outperforms state-of-the-art model-based reinforcement learning methods. And we validate MInCo's effectiveness in mitigating information conflicts through visualization.
    \vspace{-5pt}
\end{itemize}
% We evaluate MInCo on the  Deepmind Control Suite, distracted version of DeepMind Control suite (DMC) \cite{tassa2018deepmind}, which replaces the original clean backgrounds with in-the-wild videos, and Realistic Maniskill, which .
% Concretely, we train and test MInCo as well as all existing state-of-the-art baselines on 6 robotic control tasks with visual observations.
% Results demonstrate the consistent superiority of MInCo to train robot control policies.
% By conducting a series of ablation studies, we validate the efficiency of the main components of MInCo, which further verified the benefits of conflict-free latent representations.

\section{Related Work}

\subsection{Visual Model-Based Reinforcement Learning (MBRL)}
In recent years, model-based reinforcement learning has progressed from low-dimensional state spaces \cite{DBLP:conf/icml/DeisenrothR11,DBLP:journals/corr/abs-1803-00101,DBLP:conf/nips/BuckmanHTBL18,DBLP:conf/icra/NagabandiKFL18,DBLP:conf/nips/JannerFZL19} to high-dimensional visual spaces \cite{DBLP:conf/nips/WatterSBR15,DBLP:journals/corr/abs-1803-10122,DBLP:conf/icml/HafnerLFVHLD19,DBLP:conf/iclr/HafnerLB020,DBLP:conf/iclr/HafnerL0B21,DBLP:conf/l4dc/RafailovYRF21,DBLP:conf/icml/RybkinZNDML21}.
Visual MBRL takes visual observations as input and directly learns to plan or optimize the actions for the Markovian Decision Process (MDP). 
For model learning, some works \cite{DBLP:conf/nips/FinnGL16, DBLP:journals/corr/abs-1812-00568, DBLP:conf/iclr/KaiserBMOCCEFKL20} directly learn image-based dynamics,  while others \cite{DBLP:conf/nips/WatterSBR15, DBLP:conf/icml/HafnerLFVHLD19, DBLP:conf/iclr/HafnerLB020,DBLP:conf/iclr/HafnerL0B21,DBLP:journals/corr/abs-2301-04104,DBLP:conf/l4dc/RafailovYRF21,DBLP:conf/icml/RybkinZNDML21} attempt to learn dynamics from a compact latent space.
% Typically, it learns a visual world model to represent the environmental dynamics, i.e., the transition $s_{t+1}\sim P(s_{t+1}|s_t, a_t)$, which will then be used to plan or optimize actions that maximize the cumulative rewards.
% Therefore, for visual MBRL, it remains a challenging problem to learn compact and generalizable representations from partially observed vision inputs.
% PlaNet \cite{DBLP:conf/icml/HafnerLFVHLD19} proposed the recurrent state-space model (RSSM) to impose consistency on the learned dynamics beyond the next-step prediction.
Notably, the Dreamer family \cite{DBLP:conf/iclr/HafnerLB020,DBLP:conf/iclr/HafnerL0B21,DBLP:journals/corr/abs-2301-04104} explores the idea of training on ``imagination'' by optimizing the policies using generated transition data.
However, these methods suffer from information conflicts stemming from the visual reconstruction and dynamics modeling, and hence can hardly learn robust representations, especially when faced with noisy inputs.
Some methods try to learn decoupled state representations \cite{DBLP:conf/icml/FuYAJ21,DBLP:conf/nips/Pan0WY22,DBLP:conf/icml/0001D0IZT22}.
However, they still share layers and parameters between visual reconstruction and dynamics modeling, which prevents them from fully resolving information conflicts.

\subsection{Representation Learning for RL}

Representation learning \cite{DBLP:journals/pami/BengioCV13} aims to learn compact and structural representations from high-dimensional observations, which has been demonstrated to be powerful when fully-labeled data is expensive \cite{DBLP:journals/corr/abs-1807-03748,DBLP:conf/icml/ChenK0H20,DBLP:conf/cvpr/He0WXG20,DBLP:conf/nips/GrillSATRBDPGAP20,DBLP:conf/nips/CaronMMGBJ20,DBLP:conf/cvpr/ChenH21,DBLP:conf/iccv/CaronTMJMBJ21}.
Recently, these methods have been applied to reinforcement learning to learn compact representations \cite{DBLP:conf/icml/LaskinSA20,DBLP:conf/icml/StookeLAL21,DBLP:conf/iclr/DunionMLHA23}.
Some studies have demonstrated the efficiency of representation in learning reinforcement learning using techniques such as data augmentation \cite{DBLP:conf/iclr/YaratsKF21,DBLP:conf/iclr/YaratsFLP22,DBLP:conf/nips/LaskinLSPAS20,DBLP:conf/aaai/00020CX022} and masked image modeling \cite{DBLP:conf/cvpr/HeCXLDG22, DBLP:journals/corr/abs-2203-06173,DBLP:conf/nips/YuZLLC22}.
Contrastive learning, in particular, has garnered significant attention for producing robust representations for control policies.
For example, CURL \cite{DBLP:conf/icml/LaskinSA20} directly incorporates contrastive learning regularization terms into the training objective.
TED \cite{DBLP:conf/iclr/DunionMLHA23} utilizes contrastive learning to model the disentangled temporal representations of observations.
In visual model-based reinforcement learning, most works based on contrastive learning replace the visual reconstruction loss with a contrastive learning loss to facilitate generalizable representations for dynamics modeling \cite{DBLP:conf/icra/OkadaT21,DBLP:conf/iros/OkadaT22,DBLP:conf/corl/0006CHL20,DBLP:conf/icml/DengJA22,DBLP:conf/acml/WangYWL22}.
Additionally, some works also explored the idea of introducing an additional regularization based on contrastive learning for learning control policies \cite{DBLP:conf/iros/KotbWW23}.
Though powerful itself, contrastive learning can be affected by false negative or false positive samples, a challenge that has been extensively studied in self-supervised learning \cite{DBLP:conf/iccv/WangWWTL21,DBLP:conf/iclr/ChenHTC022,DBLP:conf/nips/ChuangRL0J20,DBLP:conf/nips/KalantidisSPWL20,DBLP:conf/iclr/RobinsonCSJ21,DBLP:journals/corr/abs-2401-00165}.
Nevertheless, this problem remains under-explored and unsolved in contrastive learning for reinforcement learning \cite{DBLP:conf/icml/LaskinSA20,DBLP:conf/corl/0006CHL20,DBLP:conf/iclr/DunionMLHA23,DBLP:conf/acml/WangYWL22}.

\section{Preliminaries}
\label{sec:preliminary}

\subsection{Partially Observable Markovian Decision Process (POMDP)}

A Partially Observable Markov Decision Process (POMDP) can be defined as a tuple $(\mathcal{S}, \mathcal{A}, \mathcal{O}, \mathcal{Z}, P, P_0, r,\gamma)$, where $\mathcal{S}$ is the state space, $\mathcal{A}$ is the action space, and $\mathcal{O}$ is the observation space.  $P=p(s_{t+1}|s_t,a_t)$ represents the state transition function, and $P_0$ is the initial state distribution.
$\mathcal{Z}=p(o_{t}|s_{t})$ is the observation model, which is usually used to get the posterior state distribution conditioned on the new observation. 
$r:\mathcal{S} \times \mathcal{A}  \rightarrow \mathbb{R}$ denotes the reward function, and $\gamma \in (0,1)$ is a discount factor. The goal in a POMDP is to find a policy $\pi: \mathcal{S} \rightarrow \mathcal{A}$ that maximizes the expected cumulative reward over time: $\mathbb{E}[\sum_{t=0}^{\infty}\gamma^tr(s_t, a_t)|s_0=s]$, where $s_0 \sim P_0$, $a_t \sim \pi(s_t)$, and $s_{t+1} \sim p(s_{t+1}|s_t,a_t)$.
%\begin{equation}
%	V^\pi\left(s_0\right)=\ \mathop{\mathbb{E}}_{s_0 \sim p(s_{0}),a_t \sim \pi(a_t|s_t), s_{t+1} \sim p(s_{t+1}|s_t,a_t)}[\sum_{t=0}^{\infty}r(s_t, a_t)]
%	\label{value function}
%\end{equation}

\subsection{Visual Model-Based Reinforcement Learning}

Many visual control tasks can be formalized as a POMDP. 
In addressing visual control tasks, Dreamer \cite{DBLP:conf/iclr/HafnerLB020} is one of the most influential and representative MBRL methods and serves as the baseline for many subsequent Visual MBRL approaches.
It learns a recurrent state-space model (RSSM) \cite{DBLP:conf/icml/HafnerLFVHLD19} to tackle the challenge of partial observability. 
The RSSM mainly consists of the following components:

\begin{gather}
	\begin{aligned}
		\makebox[12em][l]{Representation model:} && &p_\theta(s_t|s_{t-1},a_{t-1},o_t) \\
		\makebox[12em][l]{Observation model:} && &q_\theta(o_t|s_t) \\
		\makebox[12em][l]{Reward model:} && &q_\theta(r_t|s_t) \\
		\makebox[12em][l]{Transition model:} && &q_\theta(s_t|s_{t-1},a_{t-1}).
		\label{eq:generative_model}
	\end{aligned}
\end{gather}

Such a formulation can be regarded as a neural initiation of POMDP, with modules formulated as learnable parameters.
Under this formulation, dynamics modeling is formulated by maximizing the evidence lower bound (ELBO) \cite{DBLP:journals/ml/JordanGJS99}.
The objective function primarily consists of three parts: (a) image reconstruction \( \mathcal{J}_{\mathrm{O}}^t \doteq \ln q(o_t|s_t) \) encouraging to learn compact visual representations that can accurately reconstruct observations; (b) reward prediction \( \mathcal{J}_{\mathrm{R}}^t \doteq \ln q(r_t|s_t) \) encouraging the learned representations to be predictive for reward information; (c) dynamics regularization directly learning environmental dynamics:

\begin{equation}
    \mathcal{J}_{\mathrm{D}}^t \doteq  -\beta \mathrm{KL} (p(s_t|s_{t-1},a_{t-1},o_t) \parallel q(s_t|s_{t-1},a_{t-1})).
    \label{eq:kl}
\end{equation}
The overall objective function can be expressed as:

\begin{equation}
	\begin{aligned}
		&\mathcal{J}_{\mathrm{Dreamer}} \doteq
		\mathbb{E}_{p}\big({\sum_t\Big(\mathcal{J}_{\mathrm{O}}^t+\mathcal{J}_{\mathrm{R}}^t+\mathcal{J}_{\mathrm{D}}^t\Big)}+\text{const}\big) .\qquad 
%		\mathcal{J}_{\mathrm{O}}^t \doteq \ln q(o_t|s_t) \\
%		&\mathcal{J}_{\mathrm{R}}^t \doteq
%		\ln q(r_t|s_t) \qquad
%		\mathcal{J}_{\mathrm{D}}^t \doteq
%		\ -\beta \mathrm{KL} (p(s_t|s_{t-1},a_{t-1},o_t) \parallel q(s_t|s_{t-1},a_{t-1})).
		\label{eq:model_loss}
	\end{aligned}
\end{equation}

The expectation is taken under the representation model and a dataset consisting of trajectories.
% DreamerV2 \cite{DBLP:conf/iclr/HafnerL0B21} made improvements to the world model in order to achieve competitive performance on Atari, but did not change the overall framework of model learning.
Based on the learned predictive dynamics model, Dreamer utilizes the actor-critic method to train policy by generating imagined transition data:
% It uses the learned dynamics model to learn the following action model and value model in the latent space:

\begin{equation}
	\begin{aligned}
		\makebox[6em][l]{Action model:} && &a_\tau \sim\ q_\phi(a_\tau|s_\tau) \\
		\makebox[6em][l]{Value model:} && &v_\psi(s_\tau) \approx \mathbb{E}{q(\cdot|s_\tau)}{\textstyle\sum_{\tau=t}^{t+H}\gamma^{\tau-t}r_\tau}.
		\label{eq:action_value_model}
	\end{aligned}
\end{equation}

Specifically, during policy learning, the environmental dynamics remains frozen.
The agent imagines the trajectory of a fixed $H$ horizon in the latent space.
The value model approximates the truncated $\lambda$-return \cite{sutton2018reinforcement} through a regression loss, while the action model learns by maximizing the sum of values over the future $H$ steps.

\subsection{SimSiam}
In unsupervised visual representation learning, Siamese networks \cite{DBLP:conf/nips/BromleyGLSS93} have become a common structure. These models learn meaningful representations by maximizing the similarity between two augmented views of the same image, thereby avoiding the issue of the output collapsing into a constant. One particularly popular method is SimSiam, which effectively learns representations without using negative sample pairs, large-batch training, or momentum encoders. The architecture is as follows: two augmented views of an image are passed through an encoder network, and on one side, a prediction network is applied while a stop-gradient is enforced on the other side. The loss function is defined as follows:

\begin{equation}
	\mathcal{L} = \frac{1}{2}\mathcal{D}(p_1, \mathtt{stopgrad}(z_2)) + \frac{1}{2}\mathcal{D}(p_2, \mathtt{stopgrad}(z_1)),
 % \label{eq:simsiam}
\end{equation}
where \( \mathcal{D}(p, z) = - \frac{p \cdot z}{\|p\|_2 \|z\|_2} \)  is the negative cosine similarity, \( p_1 \) and \( p_2 \) are predictions, and \( z_1 \) and \( z_2 \) are the encoded representations.

\begin{figure*}
	\centering
	\includegraphics[width=5in]{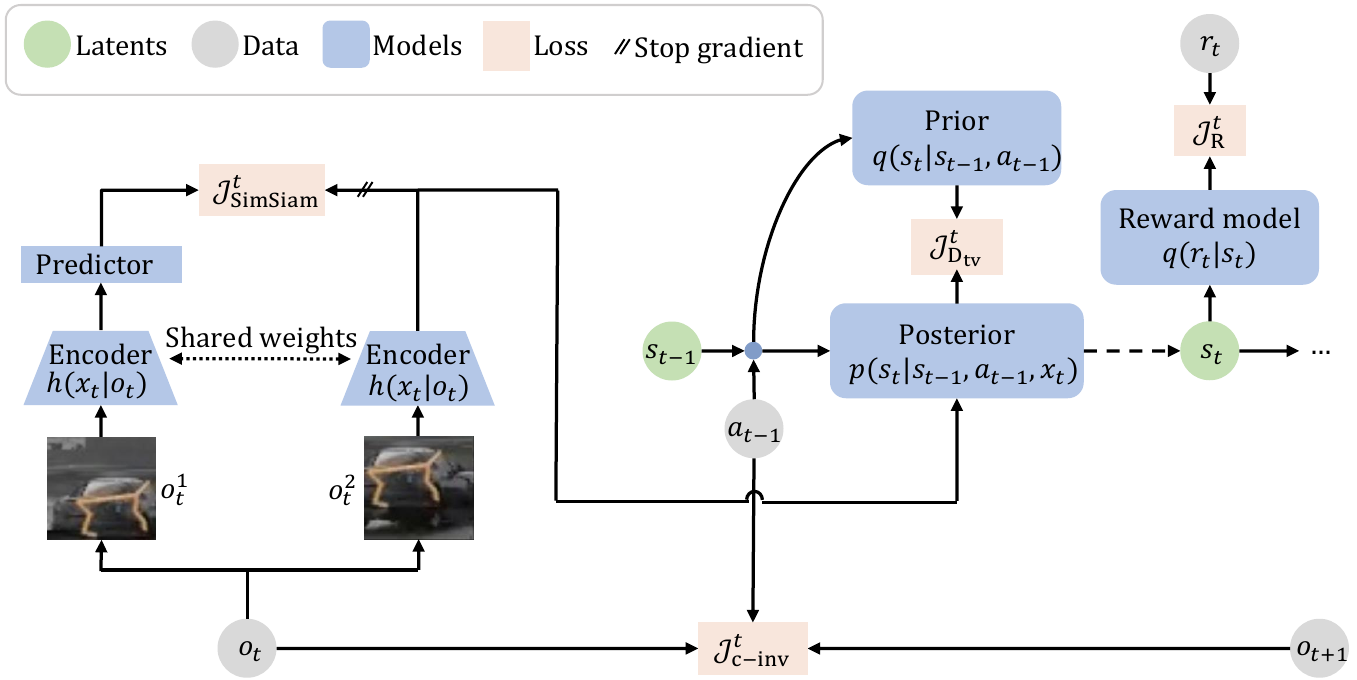}
	\caption{Overview of MInCo. MInCo learns robust representations and environment dynamics through a combination of SimSiam contrastive loss, time-varying dynamics loss, inverse dynamics loss, and the same reward reconstruction loss as Dreamer.}
    \label{fig:overview}
    \vspace{-10pt}
\end{figure*}

\section{MInCo: Mitigating Information Conflict for Visual MBRL}

In this section, we first investigate information conflicts existing in current visual MBRL (\Secref{info_confl}) algorithms.
We reveal that it originates from the visual reconstruction $\mathcal{J}_{\mathrm{O}}^t$ and dynamics modeling $\mathcal{J}_{\mathrm{D}}^t$ in the training objective (Eq. \ref{eq:model_loss}).
To mitigate such information conflicts, we propose MInCo, as shown in Fig. \ref{fig:overview}.
It consists of four models: the visual representation model, prior dynamics, posterior dynamics, and reward model.
The visual representation model is used to extract visual embeddings from the input high-dimensional visual representations $o_t$.
The resulting visual embeddings are fed into posterior dynamics to get the posterior statistical state representations, which are used for dynamics together with the prior dynamics.
Besides, to further improve decision-making performance, a reward model is introduced to regularize the learned representations to encode the reward information. 
MInCo leverages a negative-free contrastive learning algorithm, SimSiam \cite{DBLP:conf/cvpr/ChenH21}, for visual representation learning.
Besides, benefiting from the negative-free training, it solves the problem of false negatives/positives caused by InfoNCE \cite{DBLP:conf/corl/0006CHL20,DBLP:conf/acml/WangYWL22,DBLP:conf/icml/LaskinSA20} in reinforcement learning (\Secref{simsiam}).
To prevent visual representation learning from dominating the training progress, we propose the time-varying dynamics reweighting strategy, which gradually biases the training procedure towards dynamics modeling as training proceeds (\Secref{tv_dyna}). 
Finally, we use a 
cross inverse dynamics loss to facilitate the learning of controllable parts in observation (\Secref{cross_inv}).
We will introduce all technical details in this section.

\subsection{Information Conflicts in Visual MBRL}\label{info_confl}
\label{sec:info_conflict}

For visual MBRL, it is desired to learn latent state representations \( s \) that can encode the observation \( o \) sufficiently.
To achieve this, it is typical to introduce a term $\mathcal{J}_{\mathrm{O}}^t \doteq \ln q(o_t|s_t)$ in the training objective (e.g. Eq. \ref{eq:model_loss}) of visual MBRL.
From an information-theoretic perspective, we have:
\begin{proposition}\label{propos1}
	For observations \( o \) and latent states \( s \), \( \mathbb{E}[\ln(q(o|s))] \) is a lower bound of the mutual information \( I(s,o) \), i.e.,
	\begin{align}
		\mathbb{E}[\ln(q(o|s))] + h(o) &\leq \mathbb{E}[\ln\frac{q(o|s)}{p(o)}] + \mathbb{E}[\text{KL}(p(o|s) \parallel q(o|s))]  \nonumber \\ &= I(s,o),  
	\end{align}
	where \(h(o)\) is the entropy of \(O\).
\end{proposition}

% \(\mathbb{E}_{p(s,o)}\) denotes the expectation taken under the representation model and the dataset. 

Details are demonstrated in the Appendix \ref{appendix:derivation}.
Intuitively, such a visual representation learning term is actually maximizing the lower bound for the mutual information between the latent state representation $s$ and the observation $o$.
We also demonstrate that the visual contrastive objectives (e.g. InfoNCE \cite{DBLP:journals/corr/abs-1807-03748}) are also optimizing the same mutual information between $s$ and $o$ in Appendix \ref{appendix:InfoNCE}.

At the same time, it is desired that the learned latent state \( s \) can capture the information related to the decision-making, i.e., environmental dynamics (e.g. $\mathcal{J}_{\mathrm{D}}^t$ in Section \ref{sec:preliminary}). 
% Minimizing the mutual information between the latent state \( s \) and the observation \( o \) can enable \( s \) to ignore some of the information in \( o \).
From an information-theoretic perspective, we have:
\begin{proposition}\label{propos2}
	For observations \( o \), latent states \( s \) and actions \( a \), \( \mathbb{E}[\text{KL}(p(s_t|o_t,s_{t-1},a_{t-1}) \parallel q(s_t | s_{t-1},a_{t-1}))] \) is a variational upper bound of mutual information \( I(s_t,o_t | s_{t-1},a_{t-1}) \), i.e.,
	\begin{align}
		\mathbb{E}&[\text{KL}(p(s_t|o_t,s_{t-1},a_{t-1}) \parallel q(s_t|s_{t-1},a_{t-1}))] \nonumber \\ 
        &\geq I(s_t,o_t | s_{t-1},a_{t-1})
  % &\geq \mathbb{E}[\ln\frac{p(s_t|o_t,s_{t-1},a_{t-1})}{q(s_t|s_{t-1},a_{t-1})}] \nonumber \\ 
  % &-\text{KL}(p(s_t | s_{t-1},a_{t-1}) \parallel q(s_t| s_{t-1},a_{t-1}))  \nonumber \\ 
  % &=  \mathbb{E}[\ln\frac{p(s_t|o_t,s_{t-1},a_{t-1})}{p(s_t | s_{t-1},a_{t-1})}]  \nonumber \\ &\equiv  I(s_t,o_t | s_{t-1},a_{t-1})
	\end{align}
\end{proposition}
% Without loss of generality and for simplicity, we ignore the shared conditional variables (e.g. $a$ and $s$ from the last step) in Eq. \ref{eq:kl}.

The detailed derivations can be found in Appendix \ref{appendix:derivation}.
Intuitively, by maximizing the dynamics modeling loss $ \mathcal{J}_{\mathrm{D}}^t$ in Eq. \ref{eq:model_loss}, we actually hope to minimize the mutual information between the latent state representation $s$ and observation $o$.

We can see that the objective of visual representation learning $\mathcal{J}_{\mathrm{O}}^t$ and dynamics modeling $\mathcal{J}_{\mathrm{D}}^t$ are in conflict in terms of the optimization of the mutual information between the latent state $s$ and observation $o$.
They are optimizing the representations in opposite directions.
When $s_{t-1}$ and $a_{t-1}$ are given, the reconstruction term encourages $s_t$ to contain as much information from $o_t$ as possible, while the KL term constrains $s_t$ to exclude such information. As a result, the information captured in $s_t$ is encouraged to be inferred from $s_{t-1}$ and $a_{t-1}$. In a static environment, this approach does not pose significant problems because the changes in $o_t$, relative to $o_{t-1}$, are mostly task-related, and the agent can reasonably assume that these changes are driven by the action $a_{t-1}$. However, in environments with dynamic distractions, task-irrelevant background elements are also changing, and the agent mistakenly assumes that all changes in $o_t$ are caused by the action $a_{t-1}$. In such cases, the KL term indiscriminately pushes $s_t$ to exclude information from $o_t$, resulting in a conflict that hinders representation learning.
Therefore, the objective in Eq. \ref{eq:model_loss} leads to information conflicts, which harm the learning of compact and robust representations for decision-making in most visual MBRL algorithms.
In our experiments, we show that such conflicts have minor impacts on tasks with simple and static backgrounds but significantly and negatively affect tasks with noisy dynamic backgrounds with distractions.
The reason is that the learned latent state \( s \) with information conflicts fails to ignore irrelevant information, thus affecting its ability to capture the key information for decision-making.

\subsection{Negative-free Contrastive Learning for Visual MBRL}\label{simsiam}

\begin{figure}
\centering
\subfloat[]{\includegraphics[width=1in]{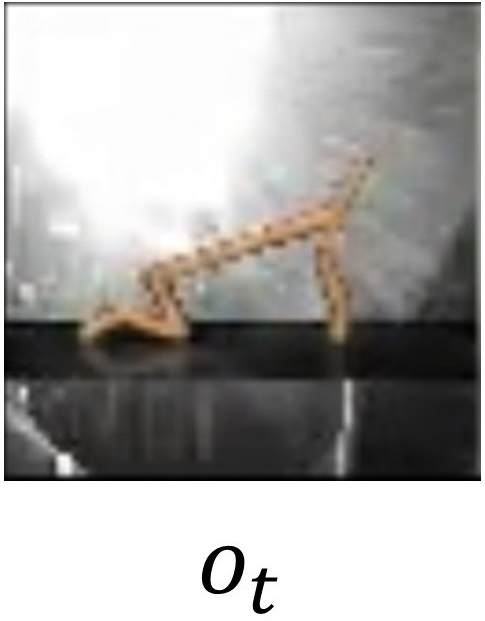}%
\label{fig1:a}}
\hfil
\subfloat[]{\includegraphics[width=1in]{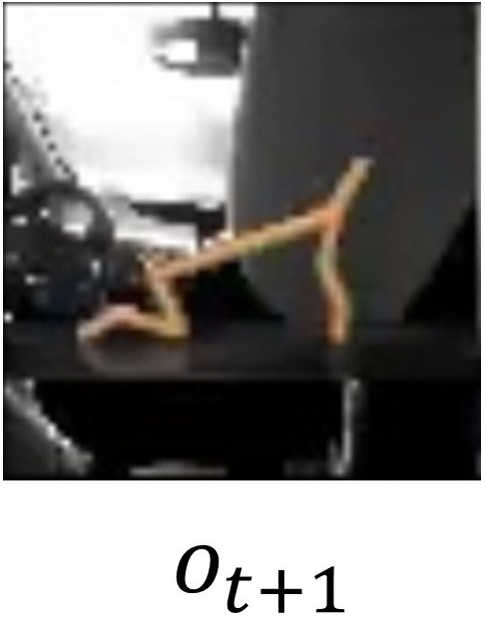}%
\label{fig1:b}}
\caption{False negatives illustration. In (a) and (b), the cheetah maintains the same posture, indicating that \( o_t \) and \( o_{t+1} \) correspond to the same latent state \( s_t \). Thus, both \( (s_t, o_t) \) and \( (s_t, o_{t+1}) \) should be considered as positive sample pairs. However, \( (s_t, o_{t+1}) \) is typically treated as a negative sample pair.}
\label{np_confusion}
\end{figure}

As discussed in Section \ref{sec:info_conflict}, information conflicts arise from visual representation learning (e.g., visual reconstruction or contrastive learning) and dynamics modeling.
Moreover, the issue of confusing samples, stemming from false positive or negative instances in contrastive learning, can adversely affect the learning of compact and robust state representations.
Such sample confusion cannot be ignored, especially at the initial stage of training.
For example, as shown in Fig. \ref{np_confusion}, the Cheetah may remain stuck in the same states for several steps due to the randomly initialized policy.
These observations, though collected at different time steps and treated as negative samples in contrastive learning, are not distinguishable from the positive ones, which results in non-negligible noise on gradients \cite{DBLP:conf/nips/ChuangRL0J20}.

% \begin{figure}
% 	\centering
% 	\begin{subfigure}[b]{0.25\textwidth}
% 		\centering
% 		\includegraphics[width=0.8in]{figures/np_confusion_a.jpg}
% 		\caption{}
% 		\label{fig1:a}
% 	\end{subfigure}%
% 	\begin{subfigure}[b]{0.25\textwidth}
% 		\centering
% 		\includegraphics[width=0.8in]{figures/np_confusion_b.jpg}
% 		\caption{}
% 		\label{fig1:b}
% 	\end{subfigure}
% 	\caption{False negatives illustration. In (a) and (b), the cheetah maintains the same posture, indicating that \( o_t \) and \( o_{t+1} \) correspond to the same latent state \( s_t \). Thus, both \( (s_t, o_t) \) and \( (s_t, o_{t+1}) \) should be considered as positive sample pairs. However, \( (s_t, o_{t+1}) \) is typically treated as a negative sample pair.}
% 	\label{np_confusion}
% \end{figure}

Essentially, these problems, particularly in the context of contrastive learning, arise from the improper selection of training samples and batches.
Therefore, to address this issue, we need to construct appropriate training batches, and then adopt conflict-free training strategies.
In this paper, we directly apply the negative-free contrastive learning algorithms, in which training batches only contain positive training samples, and hence avoid the problem of information conflicts and sample confusion.
We adopt SimSiam \cite{DBLP:conf/cvpr/ChenH21} to visual MBRL as the objective of visual representation learning due to its simplicity and high learning efficiency.

Specifically, for an observation sequence sampled from the replay buffer \( o_{1:T} \), we obtain two augmented views \( o^1_{1:T} \) and \( o^2_{1:T} \) by applying random shifts.
Noteworthily, the shifting remains consistent across different time steps in one sequence.
For each view, we obtain embeddings \( x^i_{1:T} \triangleq h(o^i_{1:T})\) using an encoder \( h \) with shared weights between the two views. 
One of the embeddings is transformed using a MLP head \( f \), i.e., \( p^1_{1:T} \triangleq f(h(o^1_{1:T})) \).
Then, we match two embeddings by maximizing their cosine similarity:

\begin{equation}
	\mathcal{D}(p^1_t, x^2_t) = \frac{p^1_t}{\left \|  p^1_t \right \|_2 } \cdot \frac{x^2_t}{\left \|  x^2_t \right \|_2 } ,	
\end{equation}

where \( 1 \leq t \leq T\), and \( \| \cdot \| \) represents the \( \mathcal{L}_2 \)-norm. A crucial component in SimSiam is the stop-gradient (\(\mathtt{stopgrad}\)) operation, which is implemented as follows:

\begin{equation}
	\mathcal{D}(p^1_t, \mathtt{stopgrad}(x^2_t)) .
\end{equation}

This means that \(x^2_t\) in this term is treated as a constant. In practice, a symmetrized loss is used:

\begin{equation}
	\mathcal{J}^t_{\mathrm{SimSiam}} = \frac{1}{2}\mathcal{D}(p^1_t, \mathtt{stopgrad}(x^2_t)) + \frac{1}{2}\mathcal{D}(p^2_t, \mathtt{stopgrad}(x^1_t)).
 \label{eq:simsiam}
\end{equation}

Eq. \ref{eq:simsiam} is defined on a single observation, and the total loss is averaged on all samples in a batch.
By using the SimSiam contrastive loss \( \mathcal{J}^t_{\mathrm{SimSiam}} \) instead of the image reconstruction loss \( \mathcal{J}_{\mathrm{O}}^t \), we mitigate information conflict while also avoiding the issue of positive-negative sample confusion.

\subsection{Time-Varying Dynamics}\label{tv_dyna}

% The source of information conflict lies in the image reconstruction loss and the dynamics loss.
In practice, we found that negative-free contrastive learning can dominate the training procedure.
By lowering its weights, the learned representations cannot sufficiently encode the observations.
To balance the training of visual representation and dynamics modeling, we propose the time-varying dynamics loss:

\begin{equation}
 \mathcal{J}_{\mathrm{D_{tv}}}^t = -\beta \mathrm{KL} (p(s_t|s_{t-1},a_{t-1},x_t) \parallel q(s_t|s_{t-1},a_{t-1})) ,
\label{eq:tvd}
\end{equation}

where:
\begin{equation}
	\beta = \min (10^{at-b},c), (a > 0, b > 0, c > 0).
 \label{eq:tvd_weight}
\end{equation}

In previous works \cite{DBLP:conf/icml/HafnerLFVHLD19,DBLP:conf/iclr/HafnerLB020,DBLP:conf/iclr/HafnerL0B21,DBLP:conf/icml/DengJA22,DBLP:conf/nips/Pan0WY22}, \( \beta \) is mostly set as a constant, such as 1.
In Eq. \ref{eq:tvd_weight}, we set \( \beta \) as an exponentially increased value according to the time step \( t \), with a maximum value of $c$.
Specifically, we adjust the initial value of \( \beta \) to the range of $10^{-5}$ to $10^{-4}$, subject to the type of tasks, when $t=0$.
This makes the ratio of \(\mathcal{J}_{\mathrm{D_{tv}}}^t\) to \( \mathcal{J}^t_{\mathrm{SimSiam}} \) close to the ratio of \( \mathcal{J}_{\mathrm{D}}^t \) to \( \mathcal{J}_{\mathrm{O}}^t \) in Eq. \ref{eq:model_loss} when \( \beta \) in \( \mathcal{J}_{\mathrm{D}}^t \) is 1. After that, \( \beta \) gradually increases as time step \( t \) increases.
As the policy gradually improves, the agent will explore different regions of the environmental dynamics, but the visual representations gradually become stable and do not change significantly anymore.

This training strategy accords with our observation that dynamics modeling is usually harder than visual representation learning.
During the learning process, visual representation learning can dominate the training at the initial stage to encourage the network to converge quickly to a stable point.
As the weight of dynamics modeling increases,  the agent is biased towards learning the reasoning ability for forward prediction, which is crucial for long-horizon decision-making.

\subsection{Cross Inverse Dynamics}\label{cross_inv}

Contrastive learning like SimSiam is designed to purely learn structural visual representations.
Nevertheless, for decision-making problems in visual MBRL, it is strongly desired that the learned visual representations focus on controllable parts of observations.
Therefore, we introduce a regularizer, namely the cross inverse dynamics loss, to facilitate the learning of controllable parts in observation.
Concretely, inspired by SPD \cite{DBLP:conf/ijcai/KimHK22}, the cross inverse dynamics loss is based on the augmented observations:

\begin{equation}
	\begin{aligned}
		\makebox[12em][l]{Cross Inverse dynamics:} && &a^1_t  =  q_\theta(h(o^1_t, o^2_{t+1})) , \\ && &a^2_t  =  q_\theta(h(o^2_t, o^1_{t+1})),
		\label{eq:cross_inv_dyn_model}
	\end{aligned}
\end{equation}
where \(h \) is vision encoder.
Specifically, we use the embedding from the first augmented view \(o^1_t \) and the embedding from the second augmented view \(o^2_{t+1} \) to recover the action between $t$ and $t+1$.
Symmetrically, we also use \(o^2_t \) and \(o^1_{t+1} \) to recover the same action.
Hence, we get the objective of cross inverse dynamics:

\begin{equation}
	\mathcal{J}_{\mathrm{c-inv}}^t = -\frac{1}{2}((a_t^1 - a_t)^2 + (a_t^2 - a_t)^2)
\end{equation}

Intuitively, this objective helps encode action-related information in learned latent state representations.

\subsection{Policy Learning}

The objective function of MInCo is a combination of $\mathcal{J}_{\mathrm{SimSiam}}^t$,
$\mathcal{J}_{\mathrm{D_{tv}}}^t$, $\mathcal{J}_{\mathrm{c-inv}}^t$, and also the inherited reward modeling loss from Dreamer $\mathcal{J}_{\mathrm{R}}^t$ \cite{DBLP:conf/iclr/HafnerLB020}:

\begin{equation}
	\begin{aligned}
		&\mathcal{J}_{\mathrm{MInCo}} \doteq
		\mathbb{E}_{p}\big({\sum_t\big(\mathcal{J}_{\mathrm{SimSiam}}^t+\mathcal{J}_{\mathrm{R}}^t+\mathcal{J}_{\mathrm{D_{tv}}}^t + \mathcal{J}_{\mathrm{c-inv}}^t\big)}\big) .
		\label{eq:minco_model_loss}
	\end{aligned}
\end{equation}

We alternate between model learning and policy learning. Model learning is performed by optimizing Eq. \ref{eq:minco_model_loss}.
The policy learning is same as previous Visual MBRL approaches \cite{DBLP:conf/iclr/HafnerLB020,DBLP:conf/iclr/HafnerL0B21}. 
Specifically, during policy learning, the environment model is fixed. We use the dynamics model to imagine trajectories of length \(H\) in the latent space, and then perform actor-critic policy learning \cite{DBLP:conf/icml/FujimotoHM18,DBLP:conf/icml/HaarnojaZAL18}. 
The critic and actor correspond to the value model and action model in Eq. \ref{eq:action_value_model}, respectively. Given a latent state, the critic is trained by predicting the truncated $\lambda$-return \cite{sutton2018reinforcement} through a regression loss, and the actor is trained to maximize the critic's prediction by performing actions. 
For more implementation details, please refer to Appendix \ref{appendix:implement}.

\begin{figure*}
\centering
\subfloat[]{\includegraphics[width=2in]{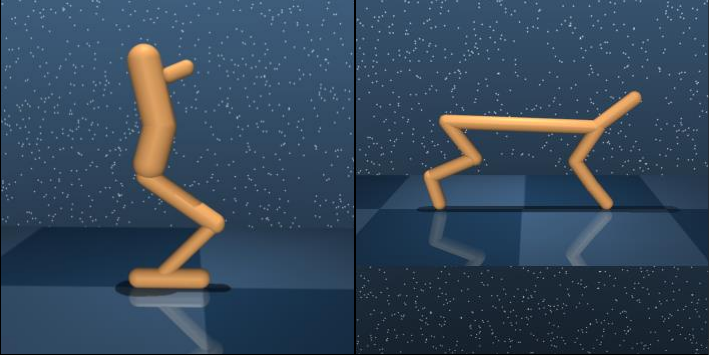}%
\label{fig:standard_dmc}}
\hspace{0.2cm}
\subfloat[]{\includegraphics[width=2in]{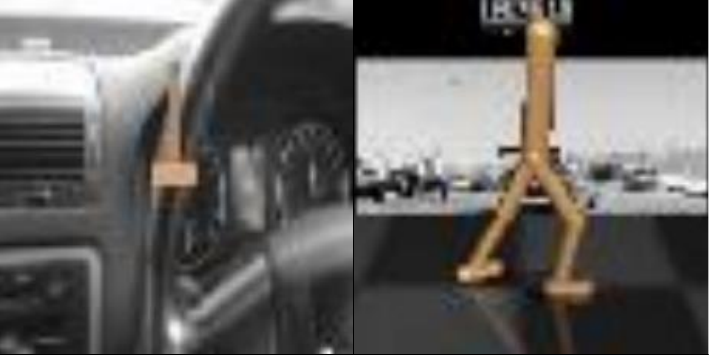}%
\label{fig:distracted_dmc}}
% \hfil
\hspace{0.2cm}
\subfloat[]{\includegraphics[width=2in]{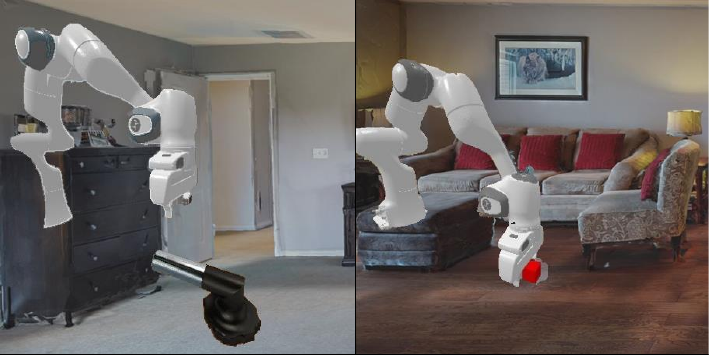}%
\label{fig:realistic_maniskill}}
\caption{Depiction of our experimental environments. (a) DeepMind Control Suite: \textit{hopper} (left) and \textit{cheetah} (right); (b) Distracted DMC: \textit{cartpole} (left) and \textit{walker} (right); (c) Realistic ManiSkill: \textit{faucet} (left) and \textit{cube} (right).}
\label{environments}
\end{figure*}

\section{Experiments}

Our experiments aim to answer the following questions: 

\begin{itemize}
    % \vspace{-5pt}
    \item Does MInCo mitigate information conflicts compared to other methods?
    \item  How is the performance of MInCo compared to the state-of-the-art visual MBRL algorithms?
    \item Is the representation learned by MInCo robust against noisy observations and distractions?
    \item What are the main contributors to the performance of MInCo?
    \vspace{-5pt}
\end{itemize}

\subsection{Experimental setup}
We evaluate our method on continuous control tasks with visual inputs. We conduct experiments in three different environments: (1) standard robotic tasks in the DeepMind Control Suite \cite{tassa2018deepmind} with static backgrounds, (2) Distracted DMC \cite{DBLP:journals/corr/abs-1811-06032}, where the static backgrounds in DMC are replaced with natural videos, and (3) the robotic arm tasks in ManiSkill2 \cite{DBLP:conf/iclr/GuXLLLMTTWYYXHC23}, which feature natural scene images of human households as backgrounds, referred to as Realistic ManiSkill \cite{DBLP:conf/nips/ZhuSG023}. The first two environments each contain 6 tasks, while the third environment contains 3 tasks, for a total of 15 tasks. Examples of the tasks are shown in the Fig. \ref{environments}.

\subsection{Baselines}
% Our experiments are conducted on distracted DeepMind Control suite (DMC) \cite{tassa2018deepmind,DBLP:journals/corr/abs-1811-06032}, of which the details can be found in Appendix \ref{appendix:implement}.
% We demonstrate the distracted DMC in Fig. \ref{environment}.
We compare MInCo with several state-of-the-art baselines that are designed to learn compact and generalizable latent representations for dynamics modeling, including:
\textbf{Dreamer} \cite{DBLP:conf/iclr/HafnerLB020}, which is a well-known visual MBRL algorithm that utilizes the learned visual dynamics to generate transitions for training policies;
\textbf{TIA} \cite{DBLP:conf/icml/FuYAJ21}, which decouples the task-relevant and task-agnostic representations for dynamics modeling;
\textbf{Iso-Dream} \cite{DBLP:conf/nips/Pan0WY22}, which utilizes inverse dynamics modeling to decouple the controllable and noncontrollable representations of dynamics, and learns policies based on the controllable part while considering the noncontrollable components;
\textbf{Denoised MDP} \cite{DBLP:conf/icml/0001D0IZT22}, which further decouples the latent representation into reward-relevant and reward-agnostic parts, and only optimizes the actions considering the reward-relevant and controllable parts;
\textbf{RePo} \cite{DBLP:conf/nips/ZhuSG023}, which encourages the learned representations to be maximally predictive of reward and dynamics and constraining the information flow from the observations; 
\textbf{HRSSM} \cite{DBLP:conf/icml/SunZ0I24}, which applies a spatio-temporal masking strategy and a bisimulation principle, combined with latent reconstruction, to capture task-specific aspects endogenous to the world model environment, thereby eliminating unnecessary information.

Additionally, a clarification regarding the baselines is necessary. Currently, there are three versions of Dreamer: DreamerV1, DreamerV2, and DreamerV3. DreamerV1 achieved strong performance on continuous tasks in DMC, while the improvements in DreamerV2 were mainly observed on Atari, as highlighted in its paper, \textit{Mastering Atari with Discrete World Models}. DreamerV3 aims to tackle tasks across multiple domains, and its performance on continuous control tasks in DMC is similar to the previous versions. In our baselines, TIA, Iso-Dream, Denoised MDP, and RePo are all based on either DreamerV1 or DreamerV2, and as such, our algorithm is also based on DreamerV1 and DreamerV2. HRSSM, on the other hand, is based on DreamerV3 and has demonstrated superior performance compared to DreamerV3. Therefore, after selecting HRSSM, we did not need to include DreamerV3 specifically for comparison. Since the overall framework of DreamerV1, DreamerV2, and DreamerV3 has not changed, our improvements are effective across all three versions.

\begin{figure}
	\centering
	\includegraphics[width=3.2in]{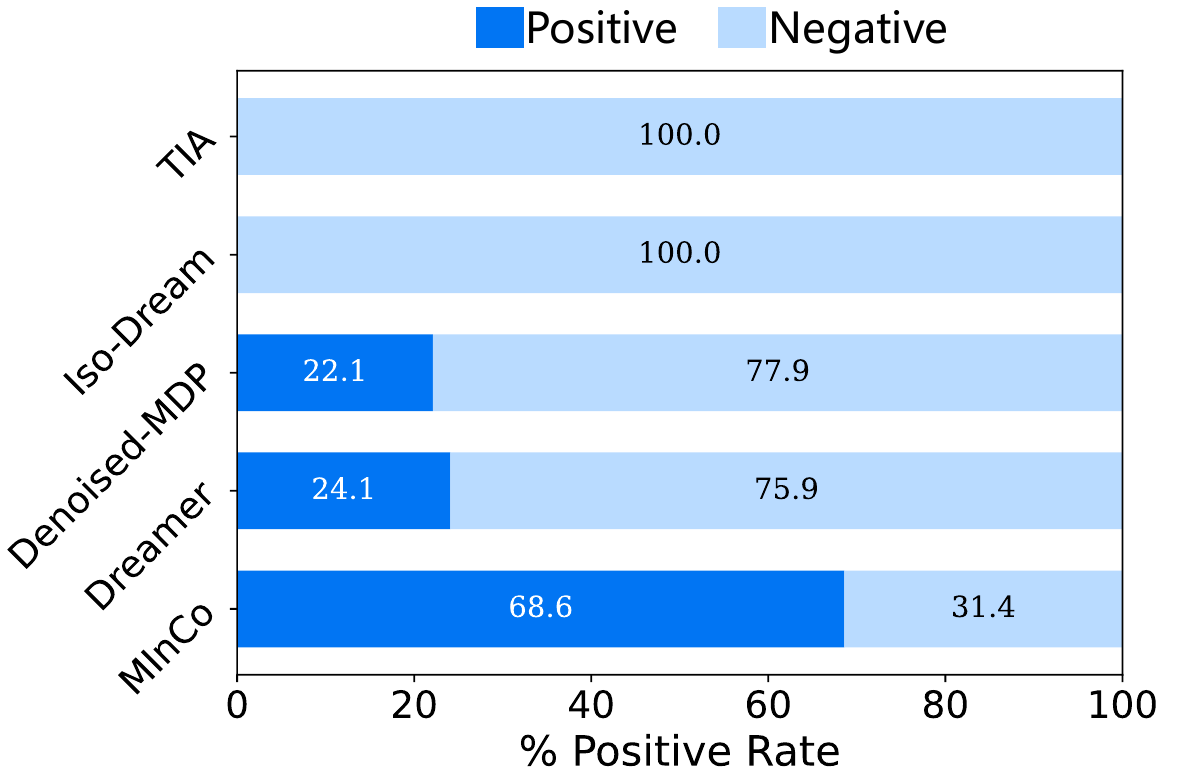}
	\caption{The normalized inner product ratio of the gradient. Compared with other methods, the ratio of our method is greater than 50\%.}
    \label{fig:grad}
    \vspace{-10pt}
\end{figure}

\subsection{Does MInCo mitigate information conflicts compared to other methods?}
There is no straightforward method to visualize information conflict. However, we use gradient information to demonstrate such conflicts. Based on our analysis, the KL term and the reconstruction term optimize the representation in opposite directions. Therefore, during training, if there is information conflict, the normalized inner product of the KL term loss gradient and the reconstruction term loss gradient with respect to the encoder should be negative most of the time. In contrast, for our method, the normalized inner product of the two gradients should be positive most of the time. We recorded the normalized inner products of the gradients for Dreamer, Iso-Dream, TIA, Denoised-MDP, and our method MInCo on the walker walk task in Distracted DMC. RePo does not have an image reconstruction loss term, and HRSSM uses latent reconstruction, so we cannot show the gradient normalized inner products for these two methods. The results are shown in Fig. \ref{fig:grad} For our method, the percentage of positive inner products is greater than 50\%, whereas for all other methods, the percentage is less than 50\%. This indicates that these methods suffer from information conflicts, while our method, MInCo, effectively mitigates this issue.

\subsection{How is the performance of MInCo compared to the state-of-the-art visual MBRL algorithms?}

\begin{figure*}
	\centering
	\includegraphics[width=5.5in]{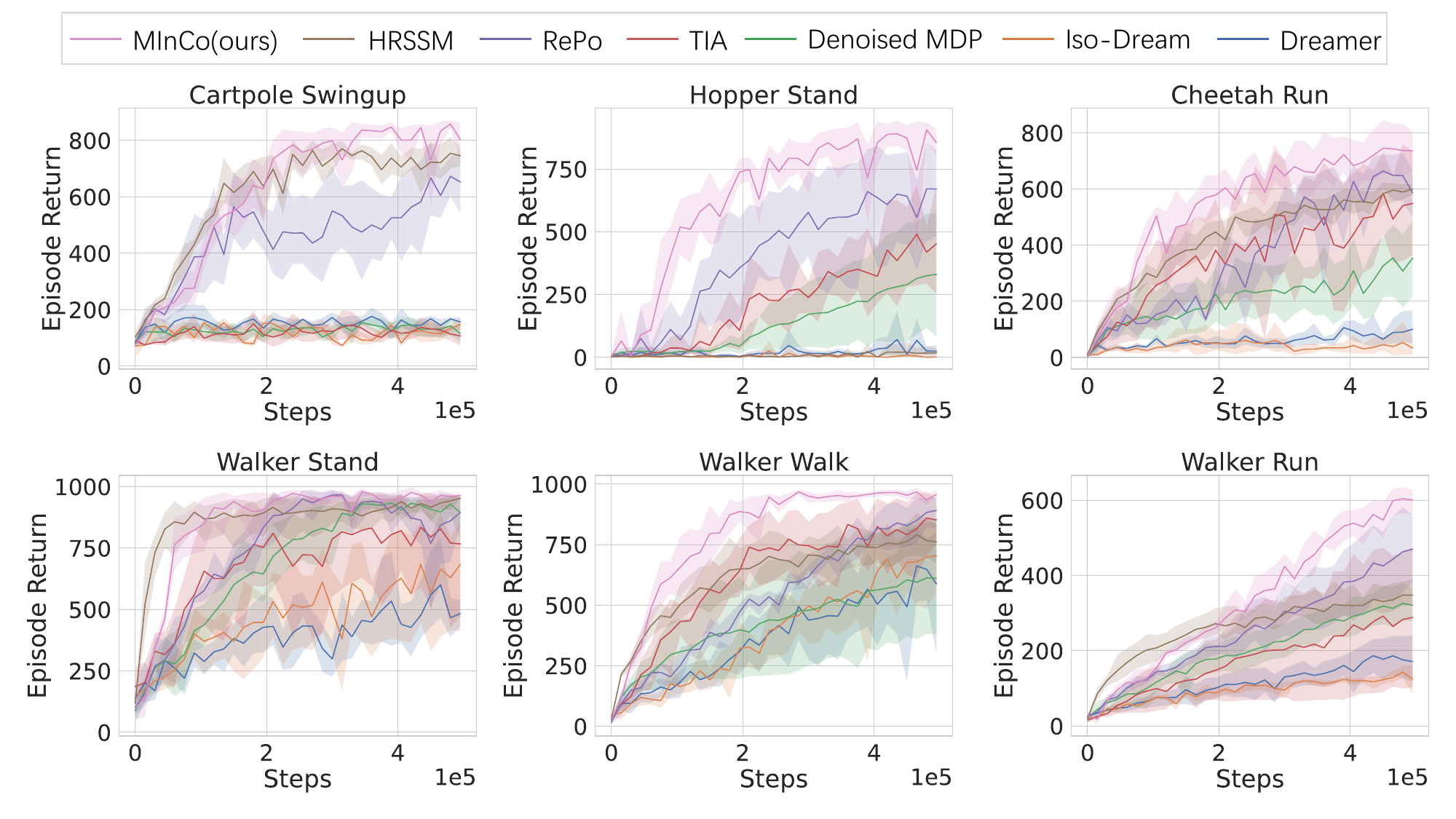}
	\caption{Experimental results on distracted DMC. These environments have dynamic background distractions. MInCo can successfully learn all of them and surpass previous visual MBRL methods in terms of learning efficiency and asymptotic performance. }
	\label{fig:performance}
\end{figure*}

\begin{figure*}
	\centering
	\includegraphics[width=5.5in]{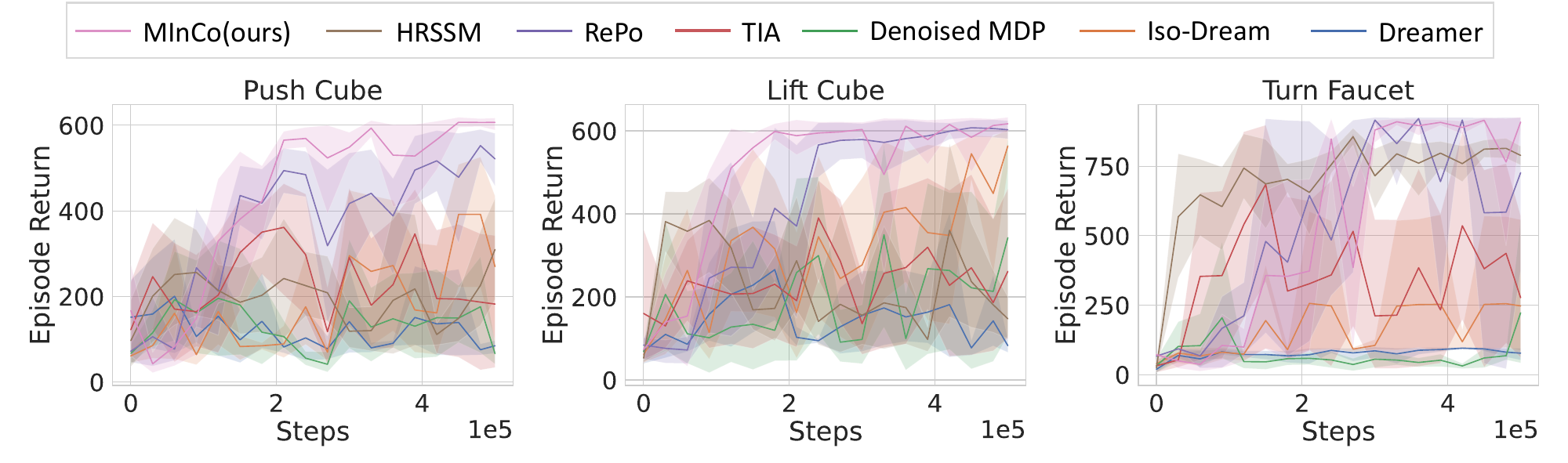}
	\caption{Experimental results on Realistic Maniskill. These environments have realistic backgrounds. MInCo can successfully learn all of them and surpass previous visual MBRL methods in terms of learning efficiency and asymptotic performance.}
	\label{fig:mani_performance}
\end{figure*}

\paragraph{MInCo consistently achieves state-of-the-art performance on tasks in the distracted DeepMind Control suite benchmark} 
Results of MInCo compared to baselines are illustrated in Fig. \ref{fig:performance}. 
We can conclude that the asymptotic performance of MInCo consistently surpasses all baselines. 
Specifically, Dreamer follows the traditional visual reconstruction loss to learn visual representations, which suffers severely from information conflicts. 
Hence, its overall performance is lower than other methods in most cases. 
Similarly, TIA, Denoised MDP, and Iso-Dream also utilize visual reconstructions for representation learning; therefore, they perform worse than RePo and MInCo due to information conflicts. 
HRSSM utilizes bisimulation and latent reconstruction, achieving strong performance on most tasks. On the Walker Stand task, it demonstrates the fastest convergence speed. However, on the Hopper Stand task, it shows almost no performance.
For RePo, although it avoids information conflicts by removing the loss of visual representations, its performance is consistently lower than MInCo, which benefits from contrastive visual representation learning.
Similarly, in the Realistic ManiSkill environment, MInCo also outperforms previous methods, as shown in Fig. \ref{fig:mani_performance}.

\paragraph{MInCo is robust against small subjects and dynamic backgrounds} Another observation is that the baselines, including Dreamer and Iso-Dream, perform relatively worse in environments where the subject of interest (e.g., the human in the walker environment or the cheetah in the cheetah environment) occupies a smaller observable area.
For example, we can see that in the Cartpole Swingup task, most of the baselines converge quickly to a much lower return compared to MInCo or HRSSM, since the car is small and will be overwhelmed by the noisy and dynamic backgrounds
By contrast, MInCo is robust against such noisy backgrounds, benefiting from the compact representation learned without information conflicts.

\begin{figure*}
	\centering
	\includegraphics[width=0.8\textwidth]{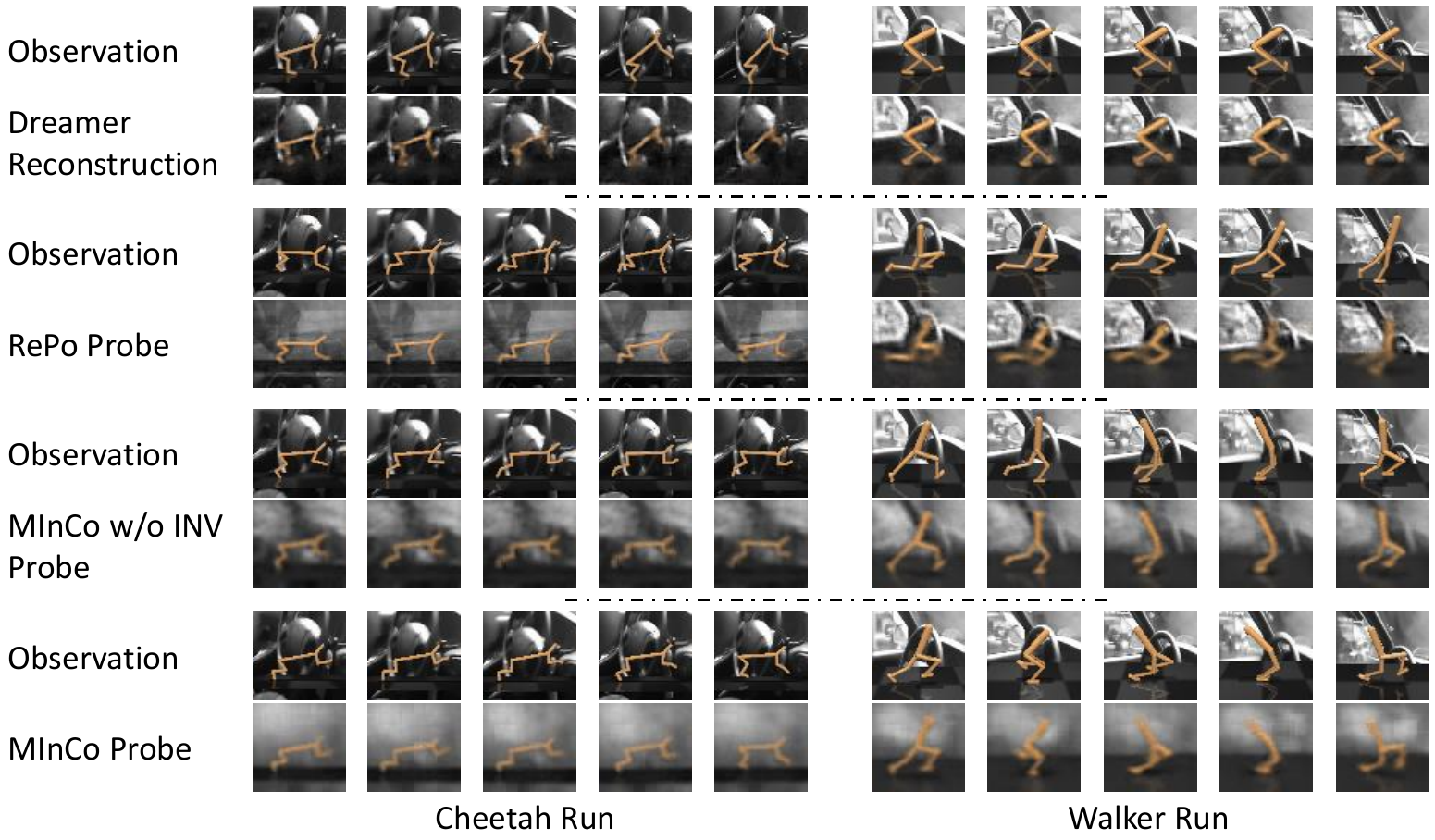}
    \captionsetup{justification=centering}
	\caption{Probing representations learned by MInCo ignore the distractions in the background.}
	\label{fig:visualization}
 \vspace{-10pt}
\end{figure*}

\paragraph{Visualization of MInCo Representations}
We also demonstrate the reconstructed visual inputs based on the latent states.
Since MInCo does not apply visual reconstruction loss for representation learning, we additionally train a decoder for reconstruction, with no gradients back-propagated to previous layers.
Results are demonstrated in Fig. \ref{fig:visualization}.
We can see that subjects in the reconstructed results of MInCo keep high motion consistency with the original inputs, while the dynamic backgrounds are ignored and filtered out.
By contrast, Dreamer gives nearly equivalent highlights on the backgrounds and subjects, and hence can only capture the motion roughly and result in less robustness against complex observations.
We also visualize the representations from RePo, which demonstrates similar performance compared to MInCo.

\begin{figure}
	\centering
	\includegraphics[width=3.2in]{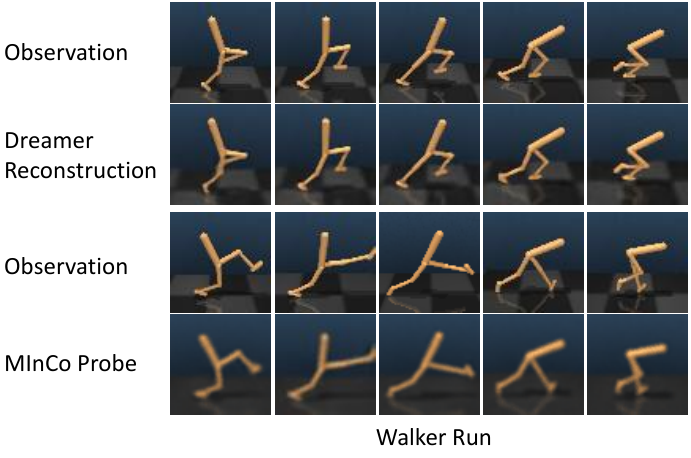}
	\caption{Probing representations learned by MInCo ignore the textures on the floor.}
    \label{fig:dmc_visualization}
    \vspace{-10pt}
\end{figure}

Similarly, we also reconstructed the visual inputs in the standard DMC environment, as shown in Fig. \ref{fig:dmc_visualization}. Compared to Dreamer, our method completely ignores the textures on the floor, focusing solely on the robot's movements.

\subsection{Is the representation learned by MInCo robust against noisy observations and distractions?}

To validate the robustness of the learned representations against noisy backgrounds, we also conduct experiments on Standard DMC.
Results are shown in Table \ref{tab:robustnessexp}. We can see that MInCo achieves comparable performance with the SOTA methods on standard DMC tasks while significantly outperforming Dreamer and RePo on distracted DMC tasks.
When transferring from DMC with clean backgrounds to DMC with dynamic and noisy distractions, MInCo has the lowest performance loss and even performs slightly better on Walker Walk and Hopper Stand.
By contrast, both Dreamer and RePo show significant performance drops with a clear margin.
It demonstrates that the learned representations by MInCo are more robust against the noise in the backgrounds.
This observation is consistent with the qualitative results as shown in Fig. \ref{fig:visualization}, where we can see that the learned representations of MInCo are sensitive to the motion of the subjects while ignoring irrelevant information in the backgrounds.

Additionally, the significant performance difference exhibited by Dreamer on standard DMC and Distracted DMC validates our assertion in \Secref{info_confl}, that information conflicts have minor impacts on tasks with simple and static backgrounds but significantly and negatively affect tasks with noisy dynamic backgrounds with distractions.

\begin{table*}[t]
\centering
\caption{Robustness against Noisy Backgrounds and Distractions.}
% \fontsize{8}{8}\selectfont
\begin{tabular}{lcccccc}
\toprule
\multirow{2}{*}{} & \multicolumn{3}{c}{Standard DMC }     & \multicolumn{3}{c}{Distracted DMC} \\
\cmidrule(lr){2-4}\cmidrule(lr){5-7}
                  & Dreamer    & RePo   & MInCo   & Dreamer    & RePo   & MInCo   \\
\midrule
Walker Stand    & 959 &  \(\sim 970\)   &  \textbf{974}      & 493 & 894    & \textbf{964}         \\
Walker Walk     & 902 &  \(\sim 920\)   & \textbf{921}    & 588 & 891   & \textbf{957}  \\
Walker Run      & 541   & \(\sim 530\)  & \textbf{638}          & 171  & 470    & \textbf{601}  \\
Cheetah Run     & 746   & \(\sim \textbf{840}\) & 825          & 100 & 587   & \textbf{736}  \\
Cartpole Swingup  & 811 & \(\sim \textbf{860}\)   & 857          & 156 & 653   & \textbf{803} \\
Hopper Stand      & \textbf{895} &\(\sim 870\)    & 884     & 24 & 670   & \textbf{854} \\

\bottomrule\\
\multicolumn{7}{c}{ * \small NOTE: the performance of RePo on Standard DMC is cited from the figures in \cite{DBLP:conf/nips/ZhuSG023}}. \\

\end{tabular}
\label{tab:robustnessexp}
\vspace{-13pt}
\end{table*}

\subsection{What are the main contributors to the performance of MInCo?}

To investigate the main contributors to the high performance of MInCo, we conduct a series of ablation studies with several variants of MInCo:
\textbf{MInCo w/o INV} is the variant that MInCo is trained without the cross inverse dynamics loss.
\textbf{MInCo w/o SimSiam} means training without the constrastive visual representation loss.
\textbf{MInCo w/o TVD} indicates the variant where the MInCo is trained without time-varying dynamics reweighting, i.e., $\beta = \min (10^{at-b},c)$ is constant in Eq. \ref{eq:tvd}. Note that in this version, we have experimented with different weights and demonstrated the best performance.

\begin{figure}
    \centering
    \includegraphics[width=3in]{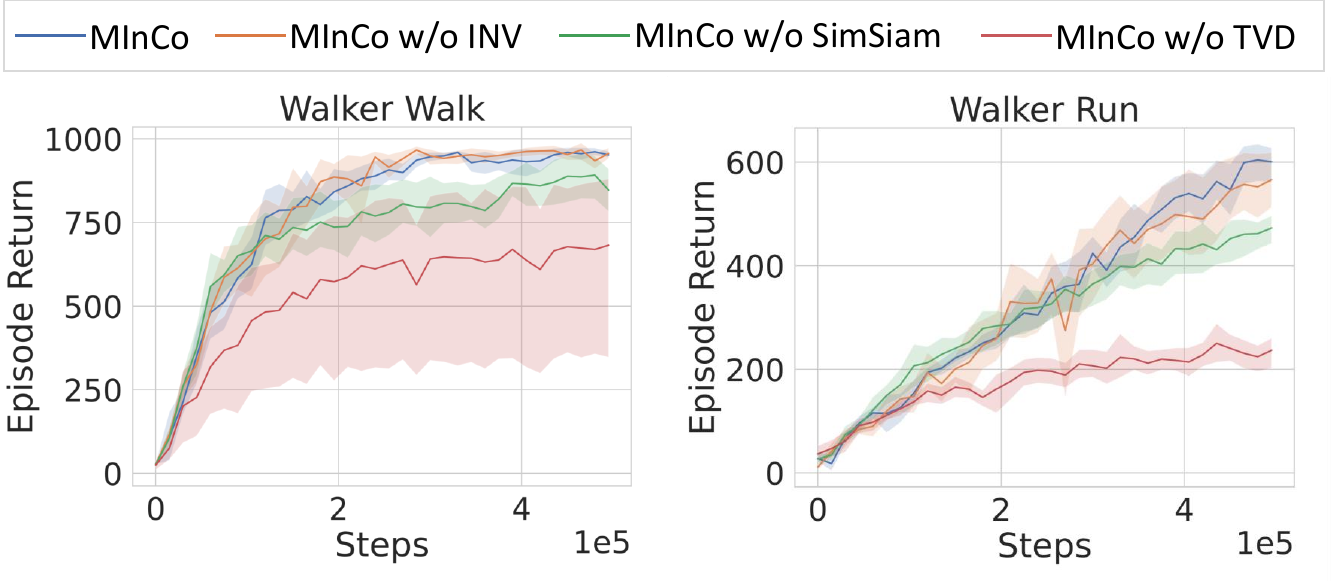}
    \caption{Ablation results of MInCo.}
    \label{fig:ablation_minco}
\end{figure}

Results are shown in Fig. \ref{fig:ablation_minco}.
We can conclude that: 
1) Time-varying dynamics reweighting is important: We can see that when training MInCo with a constant weight of the dynamics modeling, it achieves much lower performance compared to the full version due to the imbalanced learning of visual representation and dynamics; 
2) Visual representation learning contributes positively to the performance: By removing the visual representation learning, the model is trained only with dynamics modeling and reward modeling. Though avoiding information conflicts, it achieves lower performance due to the lack of supervision from observations;
3) Cross inverse dynamics can help improve the performance slightly: When using cross inverse dynamics, the learned representation contains richer information on decision-making. It helps further improve the performance slightly, especially in hard tasks.

\begin{figure*}[!t]
\centering
\subfloat[]{\includegraphics[width=3in]{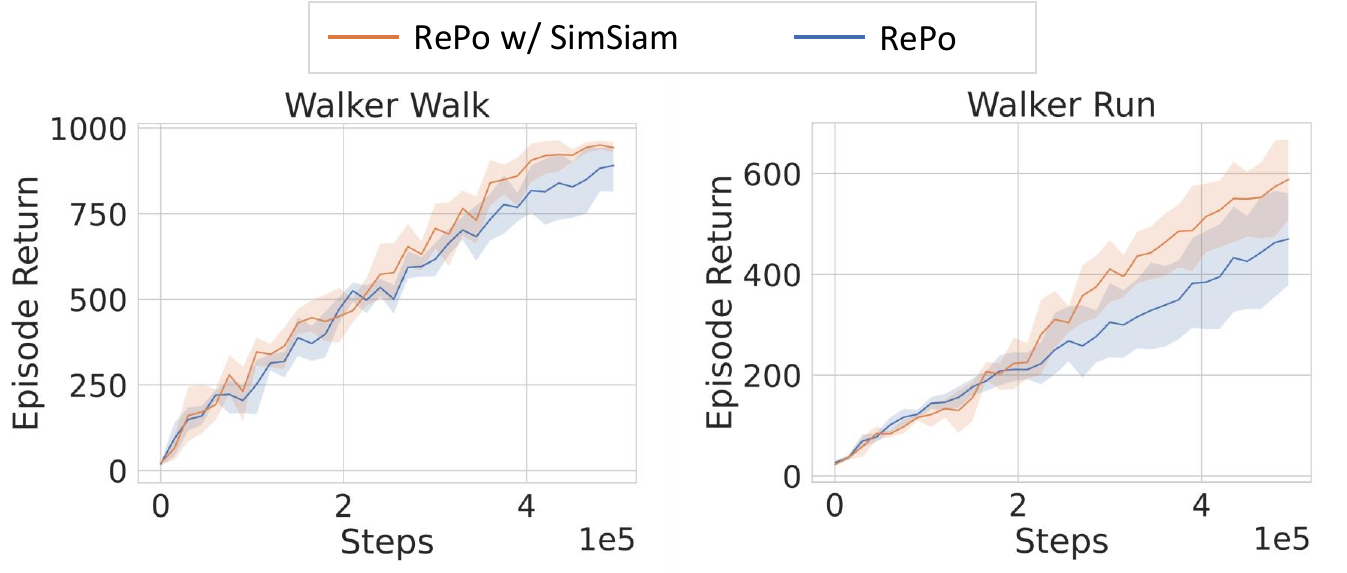}%
% \caption{RePo vs RePo with SimSiam}
\label{fig:ablatio_repo}}
% \hfil
\subfloat[]{\includegraphics[width=3in]{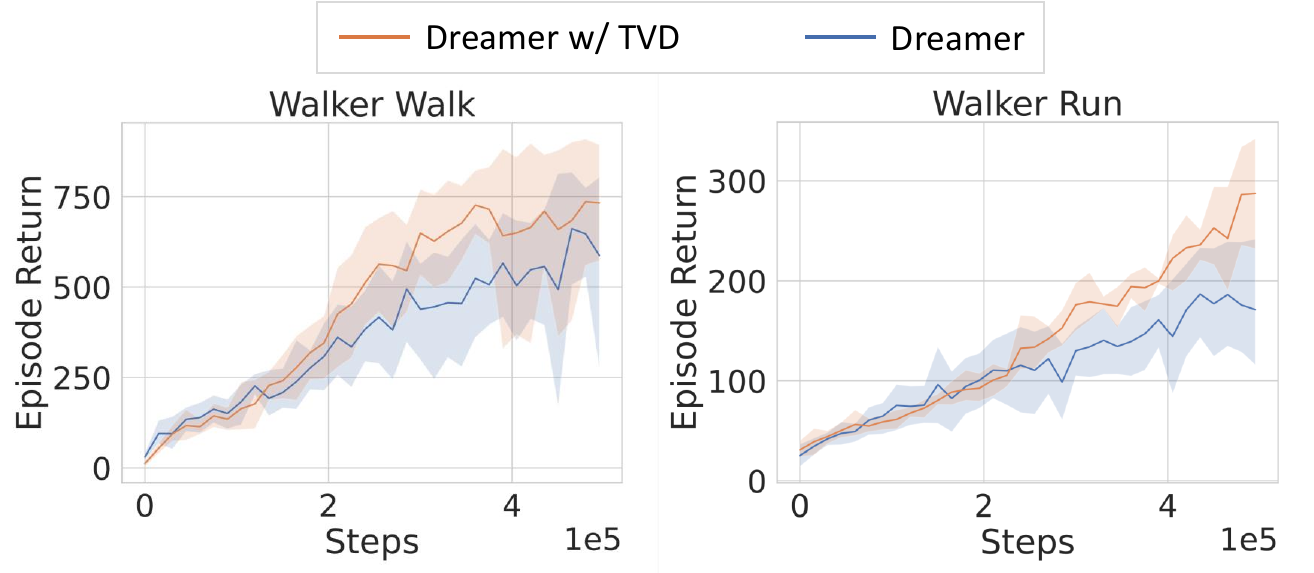}%
% \caption{Dreamer vs Dreamer with TVD}
\label{fig:ablatio_dreamer}}
\caption{Additional ablation experiments. (a) RePo vs RePo with SimSiam. (b) Dreamer vs Dreamer with TVD.}
\label{additional_ablation}
\end{figure*}

To further validate the contributors of MInCo, we have conducted two additional experiments.
First, to validate the efficiency of SimSiam, we directly add it to RePo, which only trains the dynamics modeling without explicit visual representation learning objectives.
Results are shown in Fig. \ref{fig:ablatio_repo}.
Benefiting from the additional negative-free contrastive objective of SimSiam, we can see that the performance improves consistently, which further verifies the positive impact of visual representation learning.
Second, to further validate the effectiveness of time-varying dynamics reweighting, we directly integrate it with Dreamer, which also contains objectives of both visual representation learning and dynamics modeling.
Results are shown in Fig. \ref{fig:ablatio_dreamer}
We can see that with the help of time-varying reweighting, it achieves higher performance with a clear margin. 

% We have conducted a series of ablation studies to investigate the impact of contributors on the performance of MInCo.
% Details can be found in Appendix \ref{appendix:visualization}.
From this series of ablation experiments, we can conclude that the main conclusions are:
1) Time-varying dynamics reweighting is crucial to balance the learning of visual representation and dynamics and improves the performance significantly; 
2) Visual representation learning contributes positively to the performance, which has been verified both on MInCo and RePo;
3) Cross inverse dynamics improves performance in hard tasks. However, in some simple tasks, the improvement is not statistically significant.

\subsection{Hyperparameters ablation experiments}
\begin{figure*}[!t]
\centering
\subfloat[]{\includegraphics[width=2in]{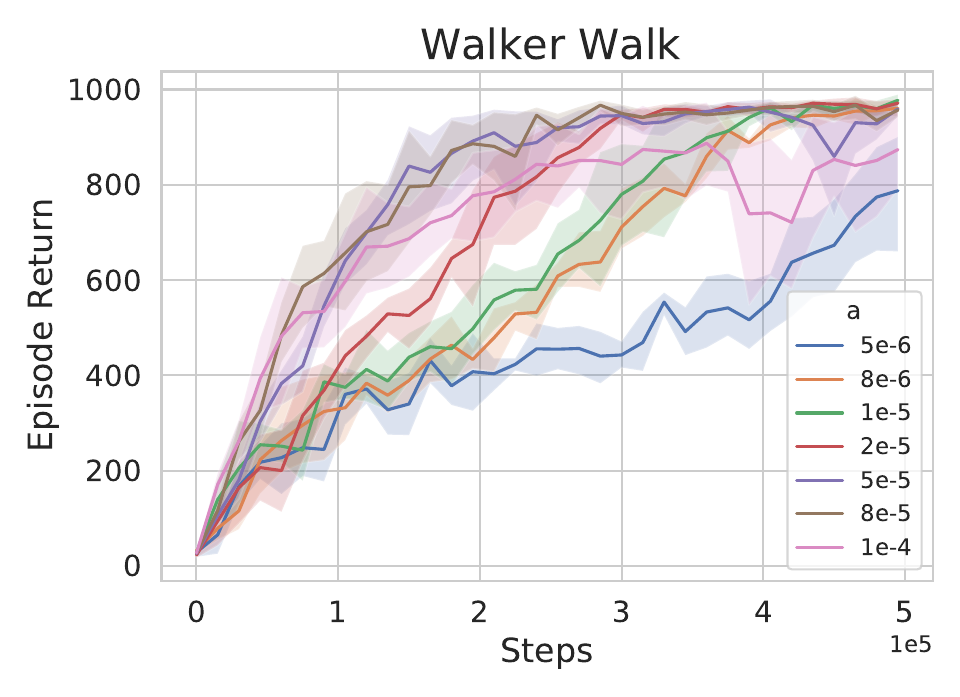}%
\label{fig:hyper_ablation_a}}
% \hfil
\subfloat[]{\includegraphics[width=2in]{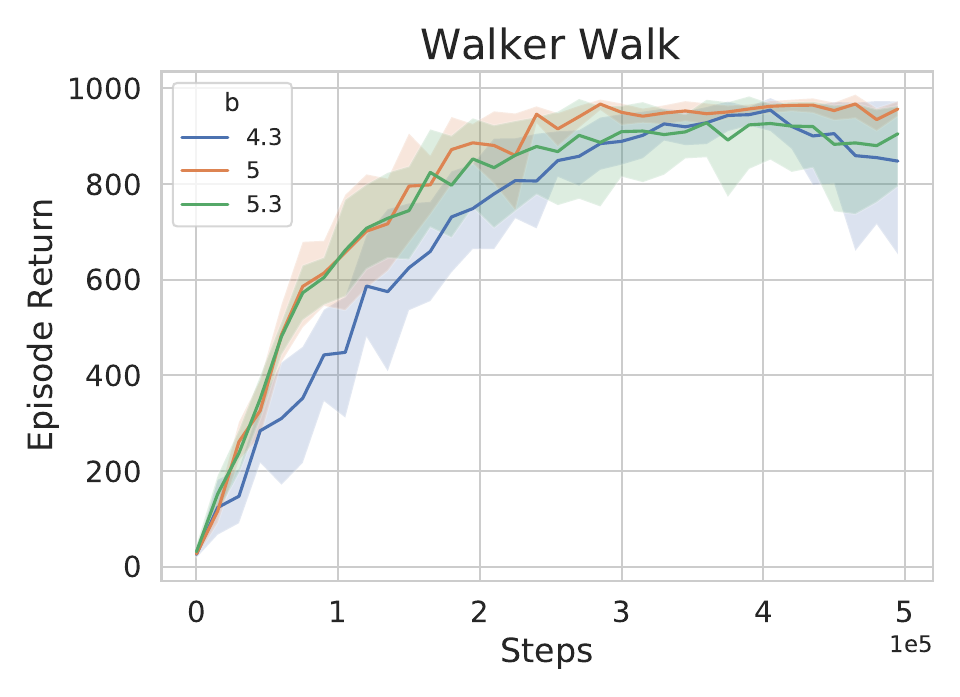}%
\label{fig:hyper_ablation_b}}
% \hfil
\subfloat[]{\includegraphics[width=2in]{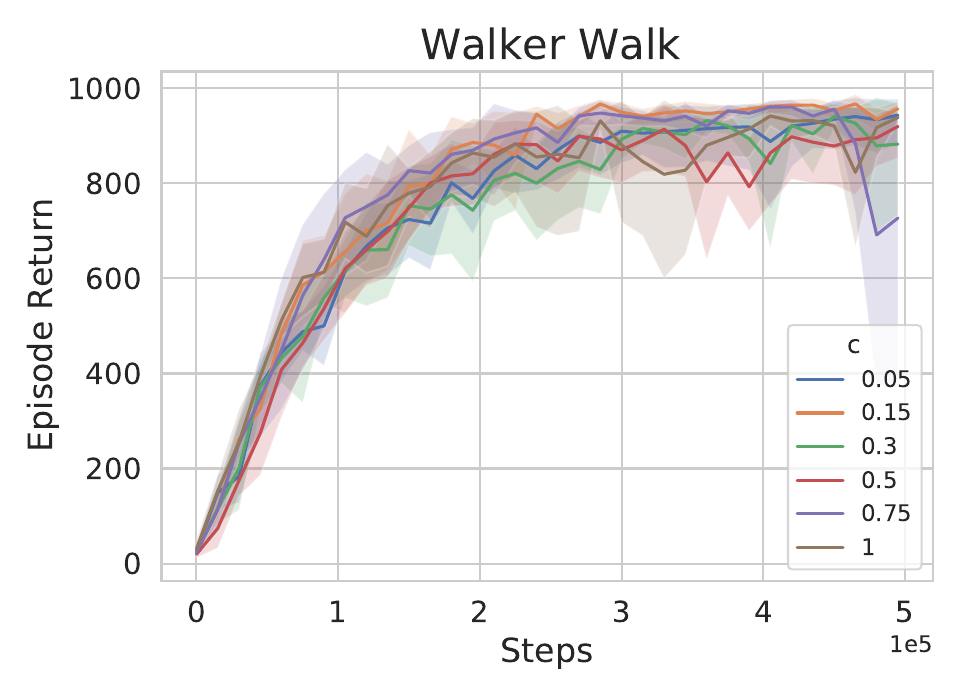}%
\label{fig:hyper_ablation_c}}
\caption{Results of hyperparameters ablation experiments. (a) Ablation results of hyperparameter \(a\). (b) Ablation results of hyperparameters \(b\). (c) Ablation results of hyperparameter \(c\).}
\label{hyper_ablation}
\end{figure*}

We also conduct ablation experiments to determine the effect of the hyperparameters in time-varying dynamics, specifically $a$, $b$, and $c$ in Eq. \ref{eq:tvd_weight}. We use the Walker Walk task of Distracted DMC as an example, and the results are shown in Fig. \ref{hyper_ablation}. We observe that as $a$ increases, the convergence speed of the algorithm significantly improves. However, when $a$ is too large, it affects the algorithm's asymptotic performance. $b$ influences the initial value of the TVD weight $\beta$, and either too large or too small of a value will impact the asymptotic performance. $c$ determines the final value of the TVD weight. When $c$ is too small, $\beta$ reaches its maximum value too quickly, which affects the convergence speed. On the other hand, if $c$ is too large, the final value of $\beta$ becomes too high, affecting the algorithm's asymptotic performance. During the experiments, we found that the value of $a$ is relatively easy to determine. For simpler tasks, such as tasks in Distracted DMC (except Walker Run), we set $a$ to $8e{-5}$, whereas for more challenging tasks, such as Walker Run in Distracted DMC and Realistic ManiSkill, we set $a$ to a smaller value of $8e{-6}$. For more details on hyperparameter settings, please refer to the appendix. \ref{appendix:implement}

\section{Conclusion}
% \vspace{-10pt}
In this paper, we reveal the problem of information conflicts, which widely exists in modern visual model-based reinforcement learning (MBRL).
It harms the learned representations to be less robust against noisy observations.
To alleviate such information conflicts, we propose MInCo, by leveraging negative-free contrastive learning and time-varying dynamics.
As a result, it can learn robust and compact representations for decision-making, despite the dynamic distractions in the backgrounds.
Our experiments demonstrate that MInCo outperforms state-of-the-art MBRL algorithms by a clear margin on the distracted DeepMind Control suite. 
Qualitative results further verify that the learned representations from MInCo are sensitive to subject motions while ignoring the background noise.
For our future work, we will adapt MInCo to more robot control tasks, including the deployment on real-robot platforms.
Besides, goal-conditioned tasks are important and ubiquitous.
Therefore, we will develop the goal-conditioned version of MInCo to solve goal-conditioned tasks.

\section*{Acknowledgments}
This work was supported in part by NSFC under grant No.62125305, No. U23A20339, No.62088102, No. 62203348.

{\appendices
\section{Proofs for Propositions 1 and 2}
\label{appendix:derivation}

The complete derivation of Proposition \ref{propos1} is as follows:
\textit{Proof.} 
\begin{equation*}
	\begin{aligned} 
		I(s,o) &\equiv \mathbb{E}[\ln\frac{p(o|s)}{p(o)}] \\
		&= \mathbb{E}[\ln\frac{q(o|s)p(o|s)}{p(o)q(o|s)}] \\
		&= \mathbb{E}[\ln\frac{q(o|s)}{p(o)}] + \mathbb{E}[\ln\frac{p(o|s)}{q(o|s)}] \\
		&= \mathbb{E}[\ln\frac{q(o|s)}{p(o)}] + \mathbb{E}[\text{KL}(p(o|s) \parallel q(o|s))] \\ 
		&\geq \mathbb{E}[\ln\frac{q(o|s)}{p(o)}] \\
		&= \mathbb{E}[\ln(q(o|s))] + \mathbb{E}[-\ln p(o)] \\
		&= \mathbb{E}[\ln(q(o|s))] + h(o), \\
	\end{aligned}
\end{equation*}
where \(h(o)\) is the entropy of \(o\). The terms in the loss function in Eq. \ref{eq:model_loss} are all summed over the time step \( t \). For convenience, we directly omit \( t \), indicating the derivation for a single time step, as the summation does not affect the result.

The complete derivation of Proposition \ref{propos2} is as follows:
% \textit{Proof.}
% \begin{equation*}
% 	\begin{aligned}
% 		& I(s_t,o_t | s_{t-1},a_{t-1}) \\ 
%   &\equiv \mathbb{E}[\ln\frac{p(s_t|o_t,s_{t-1},a_{t-1})}{p(s_t | s_{t-1},a_{t-1})}] \\
% 		&= \mathbb{E}[\ln\frac{p(s_t|o_t,s_{t-1},a_{t-1})q(s_t| s_{t-1},a_{t-1})}{q(s_t| s_{t-1},a_{t-1})p(s_t | s_{t-1},a_{t-1})}] \\
% 		&= \mathbb{E}[\ln\frac{p(s_t|o_t,s_{t-1},a_{t-1})}{q(s_t| s_{t-1},a_{t-1})}] + \mathbb{E}[\ln\frac{q(s_t| s_{t-1},a_{t-1})}{p(s_t | s_{t-1},a_{t-1})}] \\
% 		&= \mathbb{E}[\ln\frac{p(s_t|o_t,s_{t-1},a_{t-1})}{q(s_t| s_{t-1},a_{t-1})}] \\ 
%   &- \mathbb{E}[\text{KL}(p(s_t | s_{t-1},a_{t-1}) \parallel q(s_t| s_{t-1},a_{t-1}))] \\
% 		&= \mathbb{E}[\ln\frac{p(s_t|o_t,s_{t-1},a_{t-1})}{q(s_t| s_{t-1},a_{t-1})}] \\
%   &-\text{KL}(p(s_t | s_{t-1},a_{t-1}) \parallel q(s_t| s_{t-1},a_{t-1})) \\
% 		&\leq \mathbb{E}[\text{KL}(p(s_t|o_t,s_{t-1},a_{t-1}) \parallel q(s_t| s_{t-1},a_{t-1}))] .
% 	\end{aligned}
% \end{equation*}
\begin{align*}
I&(s_t, o_t \mid s_{t-1}, a_{t-1}) \\
&= \mathbb{E}_{p(s_t, o_t \mid s_{t-1}, a_{t-1})}
\left[\ln \frac{p(s_t \mid o_t, s_{t-1}, a_{t-1})}{p(s_t \mid s_{t-1}, a_{t-1})}\right] \\
&= \mathbb{E}
\left[
\ln \frac{p(s_t \mid o_t, s_{t-1}, a_{t-1})}{q(s_t \mid s_{t-1}, a_{t-1})}
- \ln \frac{p(s_t \mid s_{t-1}, a_{t-1})}{q(s_t \mid s_{t-1}, a_{t-1})}
\right] \\
% &= \mathbb{E}_{p(s_t, o_t \mid s_{t-1}, a_{t-1})}
% \ln \frac{p(s_t \mid o_t, s_{t-1}, a_{t-1})}{q(s_t \mid s_{t-1}, a_{t-1})} \\
% &\quad - \mathbb{E}_{p(s_t \mid s_{t-1}, a_{t-1})}
% \ln \frac{p(s_t \mid s_{t-1}, a_{t-1})}{q(s_t \mid s_{t-1}, a_{t-1})} \\
&= \mathbb{E}_{p(o_t \mid s_{t-1}, a_{t-1})}
\left[\mathrm{KL}\left(p(s_t \mid o_t, s_{t-1}, a_{t-1}) \,\|\, q(s_t \mid s_{t-1}, a_{t-1})\right)\right] \\
&\quad - \mathrm{KL}\left(p(s_t \mid s_{t-1}, a_{t-1}) \,\|\, q(s_t \mid s_{t-1}, a_{t-1})\right) \\
&\le \mathbb{E}_{p(o_t \mid s_{t-1}, a_{t-1})}
\mathrm{KL}\left(p(s_t \mid o_t, s_{t-1}, a_{t-1}) \,\|\, q(s_t \mid s_{t-1}, a_{t-1})\right)
\end{align*}

For further discussions on the upper and lower bounds of mutual information, please refer to \cite{DBLP:conf/icml/PooleOOAT19}.

\section{InfoNCE and Mutual Information}\label{appendix:InfoNCE}

The InfoNCE loss is a lower bound of mutual information, and this bound becomes tighter as the number of negative samples increases. Here, we provide a simple derivation; for a more detailed discussion, please refer to \cite{DBLP:journals/corr/abs-1807-03748}. For observation \(o_t\) and latent state \(s_t\), the InfoNCE loss can be defined as:

\begin{equation*}
	\mathcal{L}_{\text{InfoNCE}} = - \mathbb{E} \left[ \log \frac{f(o_t, s_t)}{ \sum_{o_j \in O}f(o_j, s_t)} \right]
\end{equation*}

In \cite{DBLP:journals/corr/abs-1807-03748}, \(f(o_t,s_t)\) is a density ratio that preserves the mutual information between \(o_t\) and \(s_t\):

\begin{equation*}
	f(o_t, s_t) \propto \frac{p(o_t|s_t)}{p(o_t)}
\end{equation*}

Then:
\begin{equation*}
	\begin{aligned}
		\mathcal{L}_{\text{InfoNCE}} &= - \mathbb{E} \left[ \log 	\frac{\frac{p(o_t|s_t)}{p(o_t)}}{\frac{p(o_t|s_t)}{p(o_t)} + \sum_{o_j \in O_{neg}} \frac{p(o_j|s_t)}{p(o_j)}} \right] \\
		&= \mathbb{E} \log \left[ 1 + \frac{p(o_t)}{p(o_t|s_t)}\sum_{o_j \in O_{neg}}\frac{p(o_j|s_t)}{p(o_j)} \right] \\
		&\approx  \mathbb{E} \log \left[ 1+ \frac{p(o_t)}{p(o_t|s_t)}(N-1) \mathbb{E}_{o_j}\frac{p(o_j|s_t)}{p(o_j)} \right] \\
		&= \mathbb{E} \log \left[ 1+ \frac{p(o_t)}{p(o_t|s_t)}(N-1) \right] \\
		&\geq \mathbb{E} \log \left[ \frac{p(o_t)}{p(o_t|s_t)}N \right] \\
		&= -I(o_t, s_t) + \log(N)
	\end{aligned}
\end{equation*}

Therefore, \( I(o_t, s_t) \geq \log(N) - \mathcal{L}_{\text{InfoNCE}}\). That is to say, minimizing \(\mathcal{L}_{\text{InfoNCE}}\) is equivalent to maximizing the lower bound mutual information \(I(o_t, s_t)\), by ignoring the constant $\log(N)$ given a fixed batch size.

\section{Implementation Details}\label{appendix:implement}

\paragraph{Evaluation Environment}
We evaluate our method as well as baseline algorithms based on the Deepmind Control Suite, distracted version of the well-known DeepMind Control suite (DMC) \cite{tassa2018deepmind,DBLP:journals/corr/abs-1811-06032}, and Realistic Maniskill \cite{DBLP:conf/nips/ZhuSG023}
Specifically, the benchmark includes 6 continuous-control tasks, including CartPole Swingup, Hopper Stand, Walker Stand, Walker Walk, Wlaker Run, and Cheetah Run.
The distractions, i.e., natural videos, are added to the background to replace the original clean background images, as shown in Fig. \ref{environments}.
The action repeat for all tasks is set to 2.
Besides, for fairness, we run 1 million steps in each experiment and record the performance and converging curves.
To make the conclusions statistically reliable, we run MInCo as well as all baselines on each of these tasks for 5 trials with different random seeds and report the mean and standard deviation.

\paragraph{Model architecture}
Our implementation of MInCo is based on Dreamer \cite{DBLP:conf/iclr/HafnerLB020}, RePo \cite{DBLP:conf/nips/ZhuSG023}, and SimSiam \cite{DBLP:conf/cvpr/ChenH21}. The observation image size is \(64 \times 64 \times 3\). The encoder utilizes a 4-layer convolutional neural network (CNN \cite{DBLP:journals/neco/LeCunBDHHHJ89}) with channel sizes {32, 64, 128, 256}, kernel size of 4, and stride of 2, with ReLU activation. The encoder output has an embedding size of 1024. The recurrent state space model is implemented using a GRU \cite{DBLP:conf/emnlp/ChoMGBBSB14}, which transits from the last state and action to the next state. Both the prior \(q(s_t|s_{t-1},a_{t-1})\) and posterior \((p(s_t|s_{t-1},a_{t-1},x_t) \) are parameterized as Gaussian distributions. The mean and standard deviation of the Gaussian distributions are predicted by a 2-layer MLP. 
The dimensionality of the state is set to 30. Similar to previous works \cite{DBLP:conf/iclr/HafnerLB020,DBLP:conf/nips/ZhuSG023}, the inputs to the reward model, policy, and value function prediction heads consist of the stochastic states with the Gaussian distribution as well as the GRU hidden variables, which are also known as deterministic states in previous works \cite{DBLP:conf/iclr/HafnerL0B21}. 
These models are all 4-layer MLPs, with the policy outputting a squashed Gaussian distribution.
All MLP hidden units are set to 200 with ELU activation. 
Our SimSiam predictor is a two-layer MLP without normalization layers.
The activation for the first layer is ReLU, and the input is the 1024-dimensional embedding from the encoder output, with the output being the same size as the input. The cosine similarity is used to compute the similarity between input and output. The inverse dynamics model is a 3-layer MLP with 512 units per layer and ELU activation. It takes the current and next image embeddings as input and predicts the action.

\paragraph{Training}
We train the agent online, executing 100 training steps every 1000 environment steps. Each training iteration involves sampling 50 trajectories, each with a length of 50, from the replay buffer to optimize the MInCo model objective Eq. \ref{eq:minco_model_loss}. Our value function estimation adopts truncated \(\lambda\)-return with \(\lambda = 0.95\) and discount factor \(\gamma = 0.99\). We employ the Adam \cite{DBLP:journals/corr/KingmaB14} optimizer for the optimization of all components. The image encoder, recurrent state-space model, reward model, SimSiam model, and inverse dynamics model share the same learning rate of 3e-4. The policy and value function use a learning rate of 8e-5. Similar to RePo \cite{DBLP:conf/nips/ZhuSG023}, we convert the KL balancing parameter \(\alpha\) into a ratio \(r\) between prior training steps and posterior training steps, where \(\alpha = \frac{r}{r+1} \), and We use the same \(r\) as RePo. We tune the \(a\), \(b\), and \(c\) in the time-varying dynamics loss. The remaining hyperparameters are configured as detailed in Table \ref{hyper-parameters}. 
% Our code is available at \url{https://anonymous.4open.science/r/minco-7C70}.

\paragraph{Baselines}
For Iso-Dream \cite{DBLP:conf/nips/Pan0WY22}\footnote{\url{https://github.com/panmt/Iso-Dream}} , Denoised MDP \cite{DBLP:conf/icml/0001D0IZT22}\footnote{\url{https://github.com/facebookresearch/denoised_mdp/}}, TIA \cite{DBLP:conf/icml/FuYAJ21}\footnote{\url{https://github.com/kyonofx/tia}}, HRSSM \cite{DBLP:conf/icml/SunZ0I24}\footnote{\url{https://github.com/bit1029public/HRSSM}} and RePo \cite{DBLP:conf/nips/ZhuSG023}\footnote{\url{https://github.com/zchuning/repo}}, we use the official implementations and their reported hyperparameters. For Denoised MDP, which overlaps with our experiments only on Cheetah Run and Walker Walk, we performe hyperparameter tuning on the other four tasks and adopt the best results. For Dreamer \cite{DBLP:conf/iclr/HafnerLB020}\footnote{\url{https://github.com/zchuning/repo/blob/main/algorithms/repo/dreamer.py}}, we use a PyTorch implementation that has comparable performance to the official version.
For the experiments in Realistic ManiSkill, we also tuned the hyperparameters for both TIA and Denoised MDP.

The experiments in this paper were conducted on an Ubuntu 16.04 system using either a single NVIDIA GeForce GTX 1080 Ti GPU or an NVIDIA GeForce RTX 2080 Ti GPU. On the Ubuntu 18.04 system, we used a single NVIDIA GeForce RTX 3090 GPU to run the experiments.

\begin{table}
    \caption{Hyperparameters Config}
    \label{hyper-parameters}
    \centering
    \begin{tabular}{cccc}
        \toprule
         & \(a\) & \(b\) & \(c\)\\
         \midrule
        Hopper Stand & 8e-5 & 4.3 & 0.015\\
        Cheetah Run & 8e-5 & 5 & 0.007\\
        Cartpole Swingup & 8e-5 & 4 & 0.0025\\
        Walker Stand & 8e-5 & 5 & 0.15\\
        Walker Walk & 8e-5 & 5 & 0.15\\
        Walker Run & 8e-6 & 5 & 0.015\\
        \midrule
        Maniskill & 8e-6 & 4 & 0.0025 \\
        \bottomrule
    \end{tabular}
    \label{tab:my_label}
\end{table}

\bibliography{references}

% Generated by IEEEtranM.bst, version: 1.13 (2008/09/30)
\ifx\mcitethebibliography\mciteundefinedmacro
\PackageError{IEEEtranM.bst}{mciteplus.sty has not been loaded}
{This bibstyle requires the use of the mciteplus package.}\fi
\begin{mcitethebibliography}{10}
\providecommand{\url}[1]{#1}
\csname url@samestyle\endcsname
\providecommand{\newblock}{\relax}
\providecommand{\bibinfo}[2]{#2}
\providecommand{\BIBentrySTDinterwordspacing}{\spaceskip=0pt\relax}
\providecommand{\BIBentryALTinterwordstretchfactor}{4}
\providecommand{\BIBentryALTinterwordspacing}{\spaceskip=\fontdimen2\font plus
\BIBentryALTinterwordstretchfactor\fontdimen3\font minus \fontdimen4\font\relax}
\providecommand{\BIBforeignlanguage}[2]{{%
\expandafter\ifx\csname l@#1\endcsname\relax
\typeout{** WARNING: IEEEtranM.bst: No hyphenation pattern has been}%
\typeout{** loaded for the language `#1'. Using the pattern for}%
\typeout{** the default language instead.}%
\else
\language=\csname l@#1\endcsname
\fi
#2}}
\providecommand{\BIBdecl}{\relax}
\BIBdecl

\bibitem{sutton2018reinforcement}
R.~S. Sutton and A.~G. Barto, \emph{Reinforcement learning: An introduction}.\hskip 1em plus 0.5em minus 0.4em\relax MIT press, 2018\relax
\mciteBstWouldAddEndPuncttrue
\mciteSetBstMidEndSepPunct{;\space}{.}{\par\relax}\relax
\EndOfBibitem
\bibitem{8890006}
Y.~Hu, W.~Wang, H.~Liu, and L.~Liu, ``Reinforcement learning tracking control for robotic manipulator with kernel-based dynamic model,'' \emph{IEEE Transactions on Neural Networks and Learning Systems}, vol.~31, no.~9, pp. 3570--3578, 2020\relax
\mciteBstWouldAddEndPuncttrue
\mciteSetBstMidEndSepPunct{;\space}{.}{\par\relax}\relax
\EndOfBibitem
\bibitem{10684782}
C.~Ma, D.~Yang, T.~Wu, Z.~Liu, H.~Yang, X.~Chen, X.~Lan, and N.~Zheng, ``Improving offline reinforcement learning with in-sample advantage regularization for robot manipulation,'' \emph{IEEE Transactions on Neural Networks and Learning Systems}, vol.~36, no.~6, pp. 11\,215--11\,227, 2025\relax
\mciteBstWouldAddEndPuncttrue
\mciteSetBstMidEndSepPunct{;\space}{.}{\par\relax}\relax
\EndOfBibitem
\bibitem{9366328}
X.~Yang, Z.~Ji, J.~Wu, Y.-K. Lai, C.~Wei, G.~Liu, and R.~Setchi, ``Hierarchical reinforcement learning with universal policies for multistep robotic manipulation,'' \emph{IEEE Transactions on Neural Networks and Learning Systems}, vol.~33, no.~9, pp. 4727--4741, 2022\relax
\mciteBstWouldAddEndPuncttrue
\mciteSetBstMidEndSepPunct{;\space}{.}{\par\relax}\relax
\EndOfBibitem
\bibitem{9582785}
Y.~Wu, S.~Liao, X.~Liu, Z.~Li, and R.~Lu, ``Deep reinforcement learning on autonomous driving policy with auxiliary critic network,'' \emph{IEEE Transactions on Neural Networks and Learning Systems}, vol.~34, no.~7, pp. 3680--3690, 2023\relax
\mciteBstWouldAddEndPuncttrue
\mciteSetBstMidEndSepPunct{;\space}{.}{\par\relax}\relax
\EndOfBibitem
\bibitem{9478933}
L.~Zhang, R.~Zhang, T.~Wu, R.~Weng, M.~Han, and Y.~Zhao, ``Safe reinforcement learning with stability guarantee for motion planning of autonomous vehicles,'' \emph{IEEE Transactions on Neural Networks and Learning Systems}, vol.~32, no.~12, pp. 5435--5444, 2021\relax
\mciteBstWouldAddEndPuncttrue
\mciteSetBstMidEndSepPunct{;\space}{.}{\par\relax}\relax
\EndOfBibitem
\bibitem{9537641}
C.~Huang, R.~Zhang, M.~Ouyang, P.~Wei, J.~Lin, J.~Su, and L.~Lin, ``Deductive reinforcement learning for visual autonomous urban driving navigation,'' \emph{IEEE Transactions on Neural Networks and Learning Systems}, vol.~32, no.~12, pp. 5379--5391, 2021\relax
\mciteBstWouldAddEndPuncttrue
\mciteSetBstMidEndSepPunct{;\space}{.}{\par\relax}\relax
\EndOfBibitem
\bibitem{DBLP:journals/nature/MnihKSRVBGRFOPB15}
\BIBentryALTinterwordspacing\relax V.~Mnih, K.~Kavukcuoglu, D.~Silver, A.~A. Rusu, J.~Veness, M.~G. Bellemare, A.~Graves, M.~A. Riedmiller, A.~Fidjeland, G.~Ostrovski, S.~Petersen, C.~Beattie, A.~Sadik, I.~Antonoglou, H.~King, D.~Kumaran, D.~Wierstra, S.~Legg, and D.~Hassabis, ``Human-level control through deep reinforcement learning,'' \emph{Nat.}, vol. 518, no. 7540, pp. 529--533, 2015. [Online]. Available: \url{https://doi.org/10.1038/nature14236}\relax
\mciteBstWouldAddEndPunctfalse
\mciteSetBstMidEndSepPunct{~;\space}{}{\par\BIBentrySTDinterwordspacing}\relax
\EndOfBibitem
\bibitem{DBLP:journals/nature/SilverSSAHGHBLB17}
\BIBentryALTinterwordspacing\relax D.~Silver, J.~Schrittwieser, K.~Simonyan, I.~Antonoglou, A.~Huang, A.~Guez, T.~Hubert, L.~Baker, M.~Lai, A.~Bolton, Y.~Chen, T.~P. Lillicrap, F.~Hui, L.~Sifre, G.~van~den Driessche, T.~Graepel, and D.~Hassabis, ``Mastering the game of go without human knowledge,'' \emph{Nat.}, vol. 550, no. 7676, pp. 354--359, 2017. [Online]. Available: \url{https://doi.org/10.1038/nature24270}\relax
\mciteBstWouldAddEndPunctfalse
\mciteSetBstMidEndSepPunct{~;\space}{}{\par\BIBentrySTDinterwordspacing}\relax
\EndOfBibitem
\bibitem{9596578}
T.~T. Nguyen and V.~J. Reddi, ``Deep reinforcement learning for cyber security,'' \emph{IEEE Transactions on Neural Networks and Learning Systems}, vol.~34, no.~8, pp. 3779--3795, 2023\relax
\mciteBstWouldAddEndPuncttrue
\mciteSetBstMidEndSepPunct{;\space}{.}{\par\relax}\relax
\EndOfBibitem
\bibitem{DBLP:conf/iclr/HafnerLB020}
\BIBentryALTinterwordspacing\relax D.~Hafner, T.~P. Lillicrap, J.~Ba, and M.~Norouzi, ``Dream to control: Learning behaviors by latent imagination,'' in \emph{8th International Conference on Learning Representations, {ICLR} 2020, Addis Ababa, Ethiopia, April 26-30, 2020}.\hskip 1em plus 0.5em minus 0.4em\relax OpenReview.net, 2020. [Online]. Available: \url{https://openreview.net/forum?id=S1lOTC4tDS}\relax
\mciteBstWouldAddEndPunctfalse
\mciteSetBstMidEndSepPunct{~;\space}{}{\par\BIBentrySTDinterwordspacing}\relax
\EndOfBibitem
\bibitem{DBLP:conf/iclr/HafnerL0B21}
\BIBentryALTinterwordspacing\relax D.~Hafner, T.~P. Lillicrap, M.~Norouzi, and J.~Ba, ``Mastering atari with discrete world models,'' in \emph{9th International Conference on Learning Representations, {ICLR} 2021, Virtual Event, Austria, May 3-7, 2021}.\hskip 1em plus 0.5em minus 0.4em\relax OpenReview.net, 2021. [Online]. Available: \url{https://openreview.net/forum?id=0oabwyZbOu}\relax
\mciteBstWouldAddEndPunctfalse
\mciteSetBstMidEndSepPunct{~;\space}{}{\par\BIBentrySTDinterwordspacing}\relax
\EndOfBibitem
\bibitem{DBLP:journals/corr/abs-2301-04104}
\BIBentryALTinterwordspacing\relax D.~Hafner, J.~Pasukonis, J.~Ba, and T.~P. Lillicrap, ``Mastering diverse domains through world models,'' \emph{CoRR}, vol. abs/2301.04104, 2023. [Online]. Available: \url{https://doi.org/10.48550/arXiv.2301.04104}\relax
\mciteBstWouldAddEndPunctfalse
\mciteSetBstMidEndSepPunct{~;\space}{}{\par\BIBentrySTDinterwordspacing}\relax
\EndOfBibitem
\bibitem{DBLP:journals/nature/SchrittwieserAH20}
\BIBentryALTinterwordspacing\relax J.~Schrittwieser, I.~Antonoglou, T.~Hubert, K.~Simonyan, L.~Sifre, S.~Schmitt, A.~Guez, E.~Lockhart, D.~Hassabis, T.~Graepel, T.~P. Lillicrap, and D.~Silver, ``Mastering atari, go, chess and shogi by planning with a learned model,'' \emph{Nat.}, vol. 588, no. 7839, pp. 604--609, 2020. [Online]. Available: \url{https://doi.org/10.1038/s41586-020-03051-4}\relax
\mciteBstWouldAddEndPunctfalse
\mciteSetBstMidEndSepPunct{~;\space}{}{\par\BIBentrySTDinterwordspacing}\relax
\EndOfBibitem
\bibitem{DBLP:conf/nips/YeLKAG21}
\BIBentryALTinterwordspacing\relax W.~Ye, S.~Liu, T.~Kurutach, P.~Abbeel, and Y.~Gao, ``Mastering atari games with limited data,'' in \emph{Advances in Neural Information Processing Systems 34: Annual Conference on Neural Information Processing Systems 2021, NeurIPS 2021, December 6-14, 2021, virtual}, M.~Ranzato, A.~Beygelzimer, Y.~N. Dauphin, P.~Liang, and J.~W. Vaughan, Eds., 2021, pp. 25\,476--25\,488. [Online]. Available: \url{https://proceedings.neurips.cc/paper/2021/hash/d5eca8dc3820cad9fe56a3bafda65ca1-Abstract.html}\relax
\mciteBstWouldAddEndPunctfalse
\mciteSetBstMidEndSepPunct{~;\space}{}{\par\BIBentrySTDinterwordspacing}\relax
\EndOfBibitem
\bibitem{DBLP:conf/icml/HansenSW22}
\BIBentryALTinterwordspacing\relax N.~Hansen, H.~Su, and X.~Wang, ``Temporal difference learning for model predictive control,'' in \emph{International Conference on Machine Learning, {ICML} 2022, 17-23 July 2022, Baltimore, Maryland, {USA}}, ser. Proceedings of Machine Learning Research, K.~Chaudhuri, S.~Jegelka, L.~Song, C.~Szepesv{\'{a}}ri, G.~Niu, and S.~Sabato, Eds., vol. 162.\hskip 1em plus 0.5em minus 0.4em\relax {PMLR}, 2022, pp. 8387--8406. [Online]. Available: \url{https://proceedings.mlr.press/v162/hansen22a.html}\relax
\mciteBstWouldAddEndPunctfalse
\mciteSetBstMidEndSepPunct{~;\space}{}{\par\BIBentrySTDinterwordspacing}\relax
\EndOfBibitem
\bibitem{DBLP:journals/corr/abs-2310-16828}
\BIBentryALTinterwordspacing\relax ------, ``{TD-MPC2:} scalable, robust world models for continuous control,'' \emph{CoRR}, vol. abs/2310.16828, 2023. [Online]. Available: \url{https://doi.org/10.48550/arXiv.2310.16828}\relax
\mciteBstWouldAddEndPunctfalse
\mciteSetBstMidEndSepPunct{~;\space}{}{\par\BIBentrySTDinterwordspacing}\relax
\EndOfBibitem
\bibitem{DBLP:conf/iclr/YamadaPGL22}
\BIBentryALTinterwordspacing\relax J.~Yamada, K.~Pertsch, A.~Gunjal, and J.~J. Lim, ``Task-induced representation learning,'' in \emph{The Tenth International Conference on Learning Representations, {ICLR} 2022, Virtual Event, April 25-29, 2022}.\hskip 1em plus 0.5em minus 0.4em\relax OpenReview.net, 2022. [Online]. Available: \url{https://openreview.net/forum?id=OzyXtIZAzFv}\relax
\mciteBstWouldAddEndPunctfalse
\mciteSetBstMidEndSepPunct{~;\space}{}{\par\BIBentrySTDinterwordspacing}\relax
\EndOfBibitem
\bibitem{DBLP:conf/ijcai/KimHK22}
\BIBentryALTinterwordspacing\relax K.~Kim, J.~Ha, and Y.~Kim, ``Self-predictive dynamics for generalization of vision-based reinforcement learning,'' in \emph{Proceedings of the Thirty-First International Joint Conference on Artificial Intelligence, {IJCAI} 2022, Vienna, Austria, 23-29 July 2022}, L.~D. Raedt, Ed.\hskip 1em plus 0.5em minus 0.4em\relax ijcai.org, 2022, pp. 3150--3156. [Online]. Available: \url{https://doi.org/10.24963/ijcai.2022/437}\relax
\mciteBstWouldAddEndPunctfalse
\mciteSetBstMidEndSepPunct{~;\space}{}{\par\BIBentrySTDinterwordspacing}\relax
\EndOfBibitem
\bibitem{DBLP:conf/aaai/0008ZY023a}
\BIBentryALTinterwordspacing\relax Q.~Liu, Q.~Zhou, R.~Yang, and J.~Wang, ``Robust representation learning by clustering with bisimulation metrics for visual reinforcement learning with distractions,'' in \emph{Thirty-Seventh {AAAI} Conference on Artificial Intelligence, {AAAI} 2023, Thirty-Fifth Conference on Innovative Applications of Artificial Intelligence, {IAAI} 2023, Thirteenth Symposium on Educational Advances in Artificial Intelligence, {EAAI} 2023, Washington, DC, USA, February 7-14, 2023}, B.~Williams, Y.~Chen, and J.~Neville, Eds.\hskip 1em plus 0.5em minus 0.4em\relax {AAAI} Press, 2023, pp. 8843--8851. [Online]. Available: \url{https://doi.org/10.1609/aaai.v37i7.26063}\relax
\mciteBstWouldAddEndPunctfalse
\mciteSetBstMidEndSepPunct{~;\space}{}{\par\BIBentrySTDinterwordspacing}\relax
\EndOfBibitem
\bibitem{DBLP:conf/corl/0006CHL20}
\BIBentryALTinterwordspacing\relax X.~Ma, S.~Chen, D.~Hsu, and W.~S. Lee, ``Contrastive variational reinforcement learning for complex observations,'' in \emph{4th Conference on Robot Learning, CoRL 2020, 16-18 November 2020, Virtual Event / Cambridge, MA, {USA}}, ser. Proceedings of Machine Learning Research, J.~Kober, F.~Ramos, and C.~J. Tomlin, Eds., vol. 155.\hskip 1em plus 0.5em minus 0.4em\relax {PMLR}, 2020, pp. 959--972. [Online]. Available: \url{https://proceedings.mlr.press/v155/ma21a.html}\relax
\mciteBstWouldAddEndPunctfalse
\mciteSetBstMidEndSepPunct{~;\space}{}{\par\BIBentrySTDinterwordspacing}\relax
\EndOfBibitem
\bibitem{DBLP:conf/acml/WangYWL22}
\BIBentryALTinterwordspacing\relax H.~Wang, X.~Yang, Y.~Wang, and X.~Lan, ``Constrained contrastive reinforcement learning,'' in \emph{Asian Conference on Machine Learning, {ACML} 2022, 12-14 December 2022, Hyderabad, India}, ser. Proceedings of Machine Learning Research, V.~N. Balasubramanian and I.~W. Tsang, Eds., vol. 189.\hskip 1em plus 0.5em minus 0.4em\relax {PMLR}, 2022, pp. 1070--1084. [Online]. Available: \url{https://proceedings.mlr.press/v189/wang23a.html}\relax
\mciteBstWouldAddEndPunctfalse
\mciteSetBstMidEndSepPunct{~;\space}{}{\par\BIBentrySTDinterwordspacing}\relax
\EndOfBibitem
\bibitem{DBLP:conf/icra/OkadaT21}
\BIBentryALTinterwordspacing\relax M.~Okada and T.~Taniguchi, ``Dreaming: Model-based reinforcement learning by latent imagination without reconstruction,'' in \emph{{IEEE} International Conference on Robotics and Automation, {ICRA} 2021, Xi'an, China, May 30 - June 5, 2021}.\hskip 1em plus 0.5em minus 0.4em\relax {IEEE}, 2021, pp. 4209--4215. [Online]. Available: \url{https://doi.org/10.1109/ICRA48506.2021.9560734}\relax
\mciteBstWouldAddEndPunctfalse
\mciteSetBstMidEndSepPunct{~;\space}{}{\par\BIBentrySTDinterwordspacing}\relax
\EndOfBibitem
\bibitem{DBLP:conf/icml/DengJA22}
\BIBentryALTinterwordspacing\relax F.~Deng, I.~Jang, and S.~Ahn, ``Dreamerpro: Reconstruction-free model-based reinforcement learning with prototypical representations,'' in \emph{International Conference on Machine Learning, {ICML} 2022, 17-23 July 2022, Baltimore, Maryland, {USA}}, ser. Proceedings of Machine Learning Research, K.~Chaudhuri, S.~Jegelka, L.~Song, C.~Szepesv{\'{a}}ri, G.~Niu, and S.~Sabato, Eds., vol. 162.\hskip 1em plus 0.5em minus 0.4em\relax {PMLR}, 2022, pp. 4956--4975. [Online]. Available: \url{https://proceedings.mlr.press/v162/deng22a.html}\relax
\mciteBstWouldAddEndPunctfalse
\mciteSetBstMidEndSepPunct{~;\space}{}{\par\BIBentrySTDinterwordspacing}\relax
\EndOfBibitem
\bibitem{DBLP:conf/iros/OkadaT22}
\BIBentryALTinterwordspacing\relax M.~Okada and T.~Taniguchi, ``Dreamingv2: Reinforcement learning with discrete world models without reconstruction,'' in \emph{{IEEE/RSJ} International Conference on Intelligent Robots and Systems, {IROS} 2022, Kyoto, Japan, October 23-27, 2022}.\hskip 1em plus 0.5em minus 0.4em\relax {IEEE}, 2022, pp. 985--991. [Online]. Available: \url{https://doi.org/10.1109/IROS47612.2022.9981405}\relax
\mciteBstWouldAddEndPunctfalse
\mciteSetBstMidEndSepPunct{~;\space}{}{\par\BIBentrySTDinterwordspacing}\relax
\EndOfBibitem
\bibitem{DBLP:conf/nips/Pan0WY22}
\BIBentryALTinterwordspacing\relax M.~Pan, X.~Zhu, Y.~Wang, and X.~Yang, ``Iso-dream: Isolating and leveraging noncontrollable visual dynamics in world models,'' in \emph{Advances in Neural Information Processing Systems 35: Annual Conference on Neural Information Processing Systems 2022, NeurIPS 2022, New Orleans, LA, USA, November 28 - December 9, 2022}, S.~Koyejo, S.~Mohamed, A.~Agarwal, D.~Belgrave, K.~Cho, and A.~Oh, Eds., 2022. [Online]. Available: \url{http://papers.nips.cc/paper\_files/paper/2022/hash/9316769afaaeeaad42a9e3633b14e801-Abstract-Conference.html}\relax
\mciteBstWouldAddEndPunctfalse
\mciteSetBstMidEndSepPunct{~;\space}{}{\par\BIBentrySTDinterwordspacing}\relax
\EndOfBibitem
\bibitem{DBLP:conf/icml/FuYAJ21}
\BIBentryALTinterwordspacing\relax X.~Fu, G.~Yang, P.~Agrawal, and T.~S. Jaakkola, ``Learning task informed abstractions,'' in \emph{Proceedings of the 38th International Conference on Machine Learning, {ICML} 2021, 18-24 July 2021, Virtual Event}, ser. Proceedings of Machine Learning Research, M.~Meila and T.~Zhang, Eds., vol. 139.\hskip 1em plus 0.5em minus 0.4em\relax {PMLR}, 2021, pp. 3480--3491. [Online]. Available: \url{http://proceedings.mlr.press/v139/fu21b.html}\relax
\mciteBstWouldAddEndPunctfalse
\mciteSetBstMidEndSepPunct{~;\space}{}{\par\BIBentrySTDinterwordspacing}\relax
\EndOfBibitem
\bibitem{DBLP:conf/icml/0001D0IZT22}
\BIBentryALTinterwordspacing\relax T.~Wang, S.~S. Du, A.~Torralba, P.~Isola, A.~Zhang, and Y.~Tian, ``Denoised mdps: Learning world models better than the world itself,'' in \emph{International Conference on Machine Learning, {ICML} 2022, 17-23 July 2022, Baltimore, Maryland, {USA}}, ser. Proceedings of Machine Learning Research, K.~Chaudhuri, S.~Jegelka, L.~Song, C.~Szepesv{\'{a}}ri, G.~Niu, and S.~Sabato, Eds., vol. 162.\hskip 1em plus 0.5em minus 0.4em\relax {PMLR}, 2022, pp. 22\,591--22\,612. [Online]. Available: \url{https://proceedings.mlr.press/v162/wang22c.html}\relax
\mciteBstWouldAddEndPunctfalse
\mciteSetBstMidEndSepPunct{~;\space}{}{\par\BIBentrySTDinterwordspacing}\relax
\EndOfBibitem
\bibitem{DBLP:conf/icml/HafnerLFVHLD19}
\BIBentryALTinterwordspacing\relax D.~Hafner, T.~P. Lillicrap, I.~Fischer, R.~Villegas, D.~Ha, H.~Lee, and J.~Davidson, ``Learning latent dynamics for planning from pixels,'' in \emph{Proceedings of the 36th International Conference on Machine Learning, {ICML} 2019, 9-15 June 2019, Long Beach, California, {USA}}, ser. Proceedings of Machine Learning Research, K.~Chaudhuri and R.~Salakhutdinov, Eds., vol.~97.\hskip 1em plus 0.5em minus 0.4em\relax {PMLR}, 2019, pp. 2555--2565. [Online]. Available: \url{http://proceedings.mlr.press/v97/hafner19a.html}\relax
\mciteBstWouldAddEndPunctfalse
\mciteSetBstMidEndSepPunct{~;\space}{}{\par\BIBentrySTDinterwordspacing}\relax
\EndOfBibitem
\bibitem{DBLP:journals/corr/abs-1807-03748}
\BIBentryALTinterwordspacing\relax A.~van~den Oord, Y.~Li, and O.~Vinyals, ``Representation learning with contrastive predictive coding,'' \emph{CoRR}, vol. abs/1807.03748, 2018. [Online]. Available: \url{http://arxiv.org/abs/1807.03748}\relax
\mciteBstWouldAddEndPunctfalse
\mciteSetBstMidEndSepPunct{~;\space}{}{\par\BIBentrySTDinterwordspacing}\relax
\EndOfBibitem
\bibitem{DBLP:conf/icml/DeisenrothR11}
\BIBentryALTinterwordspacing\relax M.~P. Deisenroth and C.~E. Rasmussen, ``{PILCO:} {A} model-based and data-efficient approach to policy search,'' in \emph{Proceedings of the 28th International Conference on Machine Learning, {ICML} 2011, Bellevue, Washington, USA, June 28 - July 2, 2011}, L.~Getoor and T.~Scheffer, Eds.\hskip 1em plus 0.5em minus 0.4em\relax Omnipress, 2011, pp. 465--472. [Online]. Available: \url{https://icml.cc/2011/papers/323\_icmlpaper.pdf}\relax
\mciteBstWouldAddEndPunctfalse
\mciteSetBstMidEndSepPunct{~;\space}{}{\par\BIBentrySTDinterwordspacing}\relax
\EndOfBibitem
\bibitem{DBLP:journals/corr/abs-1803-00101}
\BIBentryALTinterwordspacing\relax V.~Feinberg, A.~Wan, I.~Stoica, M.~I. Jordan, J.~E. Gonzalez, and S.~Levine, ``Model-based value estimation for efficient model-free reinforcement learning,'' \emph{CoRR}, vol. abs/1803.00101, 2018. [Online]. Available: \url{http://arxiv.org/abs/1803.00101}\relax
\mciteBstWouldAddEndPunctfalse
\mciteSetBstMidEndSepPunct{~;\space}{}{\par\BIBentrySTDinterwordspacing}\relax
\EndOfBibitem
\bibitem{DBLP:conf/nips/BuckmanHTBL18}
\BIBentryALTinterwordspacing\relax J.~Buckman, D.~Hafner, G.~Tucker, E.~Brevdo, and H.~Lee, ``Sample-efficient reinforcement learning with stochastic ensemble value expansion,'' in \emph{Advances in Neural Information Processing Systems 31: Annual Conference on Neural Information Processing Systems 2018, NeurIPS 2018, December 3-8, 2018, Montr{\'{e}}al, Canada}, S.~Bengio, H.~M. Wallach, H.~Larochelle, K.~Grauman, N.~Cesa{-}Bianchi, and R.~Garnett, Eds., 2018, pp. 8234--8244. [Online]. Available: \url{https://proceedings.neurips.cc/paper/2018/hash/f02208a057804ee16ac72ff4d3cec53b-Abstract.html}\relax
\mciteBstWouldAddEndPunctfalse
\mciteSetBstMidEndSepPunct{~;\space}{}{\par\BIBentrySTDinterwordspacing}\relax
\EndOfBibitem
\bibitem{DBLP:conf/icra/NagabandiKFL18}
\BIBentryALTinterwordspacing\relax A.~Nagabandi, G.~Kahn, R.~S. Fearing, and S.~Levine, ``Neural network dynamics for model-based deep reinforcement learning with model-free fine-tuning,'' in \emph{2018 {IEEE} International Conference on Robotics and Automation, {ICRA} 2018, Brisbane, Australia, May 21-25, 2018}.\hskip 1em plus 0.5em minus 0.4em\relax {IEEE}, 2018, pp. 7559--7566. [Online]. Available: \url{https://doi.org/10.1109/ICRA.2018.8463189}\relax
\mciteBstWouldAddEndPunctfalse
\mciteSetBstMidEndSepPunct{~;\space}{}{\par\BIBentrySTDinterwordspacing}\relax
\EndOfBibitem
\bibitem{DBLP:conf/nips/JannerFZL19}
\BIBentryALTinterwordspacing\relax M.~Janner, J.~Fu, M.~Zhang, and S.~Levine, ``When to trust your model: Model-based policy optimization,'' in \emph{Advances in Neural Information Processing Systems 32: Annual Conference on Neural Information Processing Systems 2019, NeurIPS 2019, December 8-14, 2019, Vancouver, BC, Canada}, H.~M. Wallach, H.~Larochelle, A.~Beygelzimer, F.~d'Alch{\'{e}}{-}Buc, E.~B. Fox, and R.~Garnett, Eds., 2019, pp. 12\,498--12\,509. [Online]. Available: \url{https://proceedings.neurips.cc/paper/2019/hash/5faf461eff3099671ad63c6f3f094f7f-Abstract.html}\relax
\mciteBstWouldAddEndPunctfalse
\mciteSetBstMidEndSepPunct{~;\space}{}{\par\BIBentrySTDinterwordspacing}\relax
\EndOfBibitem
\bibitem{DBLP:conf/nips/WatterSBR15}
\BIBentryALTinterwordspacing\relax M.~Watter, J.~T. Springenberg, J.~Boedecker, and M.~A. Riedmiller, ``Embed to control: {A} locally linear latent dynamics model for control from raw images,'' in \emph{Advances in Neural Information Processing Systems 28: Annual Conference on Neural Information Processing Systems 2015, December 7-12, 2015, Montreal, Quebec, Canada}, C.~Cortes, N.~D. Lawrence, D.~D. Lee, M.~Sugiyama, and R.~Garnett, Eds., 2015, pp. 2746--2754. [Online]. Available: \url{https://proceedings.neurips.cc/paper/2015/hash/a1afc58c6ca9540d057299ec3016d726-Abstract.html}\relax
\mciteBstWouldAddEndPunctfalse
\mciteSetBstMidEndSepPunct{~;\space}{}{\par\BIBentrySTDinterwordspacing}\relax
\EndOfBibitem
\bibitem{DBLP:journals/corr/abs-1803-10122}
\BIBentryALTinterwordspacing\relax D.~Ha and J.~Schmidhuber, ``World models,'' \emph{CoRR}, vol. abs/1803.10122, 2018. [Online]. Available: \url{http://arxiv.org/abs/1803.10122}\relax
\mciteBstWouldAddEndPunctfalse
\mciteSetBstMidEndSepPunct{~;\space}{}{\par\BIBentrySTDinterwordspacing}\relax
\EndOfBibitem
\bibitem{DBLP:conf/l4dc/RafailovYRF21}
\BIBentryALTinterwordspacing\relax R.~Rafailov, T.~Yu, A.~Rajeswaran, and C.~Finn, ``Offline reinforcement learning from images with latent space models,'' in \emph{Proceedings of the 3rd Annual Conference on Learning for Dynamics and Control, {L4DC} 2021, 7-8 June 2021, Virtual Event, Switzerland}, ser. Proceedings of Machine Learning Research, A.~Jadbabaie, J.~Lygeros, G.~J. Pappas, P.~A. Parrilo, B.~Recht, C.~J. Tomlin, and M.~N. Zeilinger, Eds., vol. 144.\hskip 1em plus 0.5em minus 0.4em\relax {PMLR}, 2021, pp. 1154--1168. [Online]. Available: \url{http://proceedings.mlr.press/v144/rafailov21a.html}\relax
\mciteBstWouldAddEndPunctfalse
\mciteSetBstMidEndSepPunct{~;\space}{}{\par\BIBentrySTDinterwordspacing}\relax
\EndOfBibitem
\bibitem{DBLP:conf/icml/RybkinZNDML21}
\BIBentryALTinterwordspacing\relax O.~Rybkin, C.~Zhu, A.~Nagabandi, K.~Daniilidis, I.~Mordatch, and S.~Levine, ``Model-based reinforcement learning via latent-space collocation,'' in \emph{Proceedings of the 38th International Conference on Machine Learning, {ICML} 2021, 18-24 July 2021, Virtual Event}, ser. Proceedings of Machine Learning Research, M.~Meila and T.~Zhang, Eds., vol. 139.\hskip 1em plus 0.5em minus 0.4em\relax {PMLR}, 2021, pp. 9190--9201. [Online]. Available: \url{http://proceedings.mlr.press/v139/rybkin21b.html}\relax
\mciteBstWouldAddEndPunctfalse
\mciteSetBstMidEndSepPunct{~;\space}{}{\par\BIBentrySTDinterwordspacing}\relax
\EndOfBibitem
\bibitem{DBLP:conf/nips/FinnGL16}
\BIBentryALTinterwordspacing\relax C.~Finn, I.~J. Goodfellow, and S.~Levine, ``Unsupervised learning for physical interaction through video prediction,'' in \emph{Advances in Neural Information Processing Systems 29: Annual Conference on Neural Information Processing Systems 2016, December 5-10, 2016, Barcelona, Spain}, D.~D. Lee, M.~Sugiyama, U.~von Luxburg, I.~Guyon, and R.~Garnett, Eds., 2016, pp. 64--72. [Online]. Available: \url{https://proceedings.neurips.cc/paper/2016/hash/d9d4f495e875a2e075a1a4a6e1b9770f-Abstract.html}\relax
\mciteBstWouldAddEndPunctfalse
\mciteSetBstMidEndSepPunct{~;\space}{}{\par\BIBentrySTDinterwordspacing}\relax
\EndOfBibitem
\bibitem{DBLP:journals/corr/abs-1812-00568}
\BIBentryALTinterwordspacing\relax F.~Ebert, C.~Finn, S.~Dasari, A.~Xie, A.~X. Lee, and S.~Levine, ``Visual foresight: Model-based deep reinforcement learning for vision-based robotic control,'' \emph{CoRR}, vol. abs/1812.00568, 2018. [Online]. Available: \url{http://arxiv.org/abs/1812.00568}\relax
\mciteBstWouldAddEndPunctfalse
\mciteSetBstMidEndSepPunct{~;\space}{}{\par\BIBentrySTDinterwordspacing}\relax
\EndOfBibitem
\bibitem{DBLP:conf/iclr/KaiserBMOCCEFKL20}
\BIBentryALTinterwordspacing\relax L.~Kaiser, M.~Babaeizadeh, P.~Milos, B.~Osinski, R.~H. Campbell, K.~Czechowski, D.~Erhan, C.~Finn, P.~Kozakowski, S.~Levine, A.~Mohiuddin, R.~Sepassi, G.~Tucker, and H.~Michalewski, ``Model based reinforcement learning for atari,'' in \emph{8th International Conference on Learning Representations, {ICLR} 2020, Addis Ababa, Ethiopia, April 26-30, 2020}.\hskip 1em plus 0.5em minus 0.4em\relax OpenReview.net, 2020. [Online]. Available: \url{https://openreview.net/forum?id=S1xCPJHtDB}\relax
\mciteBstWouldAddEndPunctfalse
\mciteSetBstMidEndSepPunct{~;\space}{}{\par\BIBentrySTDinterwordspacing}\relax
\EndOfBibitem
\bibitem{DBLP:journals/pami/BengioCV13}
\BIBentryALTinterwordspacing\relax Y.~Bengio, A.~C. Courville, and P.~Vincent, ``Representation learning: {A} review and new perspectives,'' \emph{{IEEE} Trans. Pattern Anal. Mach. Intell.}, vol.~35, no.~8, pp. 1798--1828, 2013. [Online]. Available: \url{https://doi.org/10.1109/TPAMI.2013.50}\relax
\mciteBstWouldAddEndPunctfalse
\mciteSetBstMidEndSepPunct{~;\space}{}{\par\BIBentrySTDinterwordspacing}\relax
\EndOfBibitem
\bibitem{DBLP:conf/icml/ChenK0H20}
\BIBentryALTinterwordspacing\relax T.~Chen, S.~Kornblith, M.~Norouzi, and G.~E. Hinton, ``A simple framework for contrastive learning of visual representations,'' in \emph{Proceedings of the 37th International Conference on Machine Learning, {ICML} 2020, 13-18 July 2020, Virtual Event}, ser. Proceedings of Machine Learning Research, vol. 119.\hskip 1em plus 0.5em minus 0.4em\relax {PMLR}, 2020, pp. 1597--1607. [Online]. Available: \url{http://proceedings.mlr.press/v119/chen20j.html}\relax
\mciteBstWouldAddEndPunctfalse
\mciteSetBstMidEndSepPunct{~;\space}{}{\par\BIBentrySTDinterwordspacing}\relax
\EndOfBibitem
\bibitem{DBLP:conf/cvpr/He0WXG20}
\BIBentryALTinterwordspacing\relax K.~He, H.~Fan, Y.~Wu, S.~Xie, and R.~B. Girshick, ``Momentum contrast for unsupervised visual representation learning,'' in \emph{2020 {IEEE/CVF} Conference on Computer Vision and Pattern Recognition, {CVPR} 2020, Seattle, WA, USA, June 13-19, 2020}.\hskip 1em plus 0.5em minus 0.4em\relax Computer Vision Foundation / {IEEE}, 2020, pp. 9726--9735. [Online]. Available: \url{https://doi.org/10.1109/CVPR42600.2020.00975}\relax
\mciteBstWouldAddEndPunctfalse
\mciteSetBstMidEndSepPunct{~;\space}{}{\par\BIBentrySTDinterwordspacing}\relax
\EndOfBibitem
\bibitem{DBLP:conf/nips/GrillSATRBDPGAP20}
\BIBentryALTinterwordspacing\relax J.~Grill, F.~Strub, F.~Altch{\'{e}}, C.~Tallec, P.~H. Richemond, E.~Buchatskaya, C.~Doersch, B.~{\'{A}}. Pires, Z.~Guo, M.~G. Azar, B.~Piot, K.~Kavukcuoglu, R.~Munos, and M.~Valko, ``Bootstrap your own latent - {A} new approach to self-supervised learning,'' in \emph{Advances in Neural Information Processing Systems 33: Annual Conference on Neural Information Processing Systems 2020, NeurIPS 2020, December 6-12, 2020, virtual}, H.~Larochelle, M.~Ranzato, R.~Hadsell, M.~Balcan, and H.~Lin, Eds., 2020. [Online]. Available: \url{https://proceedings.neurips.cc/paper/2020/hash/f3ada80d5c4ee70142b17b8192b2958e-Abstract.html}\relax
\mciteBstWouldAddEndPunctfalse
\mciteSetBstMidEndSepPunct{~;\space}{}{\par\BIBentrySTDinterwordspacing}\relax
\EndOfBibitem
\bibitem{DBLP:conf/nips/CaronMMGBJ20}
\BIBentryALTinterwordspacing\relax M.~Caron, I.~Misra, J.~Mairal, P.~Goyal, P.~Bojanowski, and A.~Joulin, ``Unsupervised learning of visual features by contrasting cluster assignments,'' in \emph{Advances in Neural Information Processing Systems 33: Annual Conference on Neural Information Processing Systems 2020, NeurIPS 2020, December 6-12, 2020, virtual}, H.~Larochelle, M.~Ranzato, R.~Hadsell, M.~Balcan, and H.~Lin, Eds., 2020. [Online]. Available: \url{https://proceedings.neurips.cc/paper/2020/hash/70feb62b69f16e0238f741fab228fec2-Abstract.html}\relax
\mciteBstWouldAddEndPunctfalse
\mciteSetBstMidEndSepPunct{~;\space}{}{\par\BIBentrySTDinterwordspacing}\relax
\EndOfBibitem
\bibitem{DBLP:conf/cvpr/ChenH21}
\BIBentryALTinterwordspacing\relax X.~Chen and K.~He, ``Exploring simple siamese representation learning,'' in \emph{{IEEE} Conference on Computer Vision and Pattern Recognition, {CVPR} 2021, virtual, June 19-25, 2021}.\hskip 1em plus 0.5em minus 0.4em\relax Computer Vision Foundation / {IEEE}, 2021, pp. 15\,750--15\,758. [Online]. Available: \url{https://openaccess.thecvf.com/content/CVPR2021/html/Chen\_Exploring\_Simple\_Siamese\_Representation\_Learning\_CVPR\_2021\_paper.html}\relax
\mciteBstWouldAddEndPunctfalse
\mciteSetBstMidEndSepPunct{~;\space}{}{\par\BIBentrySTDinterwordspacing}\relax
\EndOfBibitem
\bibitem{DBLP:conf/iccv/CaronTMJMBJ21}
\BIBentryALTinterwordspacing\relax M.~Caron, H.~Touvron, I.~Misra, H.~J{\'{e}}gou, J.~Mairal, P.~Bojanowski, and A.~Joulin, ``Emerging properties in self-supervised vision transformers,'' in \emph{2021 {IEEE/CVF} International Conference on Computer Vision, {ICCV} 2021, Montreal, QC, Canada, October 10-17, 2021}.\hskip 1em plus 0.5em minus 0.4em\relax {IEEE}, 2021, pp. 9630--9640. [Online]. Available: \url{https://doi.org/10.1109/ICCV48922.2021.00951}\relax
\mciteBstWouldAddEndPunctfalse
\mciteSetBstMidEndSepPunct{~;\space}{}{\par\BIBentrySTDinterwordspacing}\relax
\EndOfBibitem
\bibitem{DBLP:conf/icml/LaskinSA20}
\BIBentryALTinterwordspacing\relax M.~Laskin, A.~Srinivas, and P.~Abbeel, ``{CURL:} contrastive unsupervised representations for reinforcement learning,'' in \emph{Proceedings of the 37th International Conference on Machine Learning, {ICML} 2020, 13-18 July 2020, Virtual Event}, ser. Proceedings of Machine Learning Research, vol. 119.\hskip 1em plus 0.5em minus 0.4em\relax {PMLR}, 2020, pp. 5639--5650. [Online]. Available: \url{http://proceedings.mlr.press/v119/laskin20a.html}\relax
\mciteBstWouldAddEndPunctfalse
\mciteSetBstMidEndSepPunct{~;\space}{}{\par\BIBentrySTDinterwordspacing}\relax
\EndOfBibitem
\bibitem{DBLP:conf/icml/StookeLAL21}
\BIBentryALTinterwordspacing\relax A.~Stooke, K.~Lee, P.~Abbeel, and M.~Laskin, ``Decoupling representation learning from reinforcement learning,'' in \emph{Proceedings of the 38th International Conference on Machine Learning, {ICML} 2021, 18-24 July 2021, Virtual Event}, ser. Proceedings of Machine Learning Research, M.~Meila and T.~Zhang, Eds., vol. 139.\hskip 1em plus 0.5em minus 0.4em\relax {PMLR}, 2021, pp. 9870--9879. [Online]. Available: \url{http://proceedings.mlr.press/v139/stooke21a.html}\relax
\mciteBstWouldAddEndPunctfalse
\mciteSetBstMidEndSepPunct{~;\space}{}{\par\BIBentrySTDinterwordspacing}\relax
\EndOfBibitem
\bibitem{DBLP:conf/iclr/DunionMLHA23}
\BIBentryALTinterwordspacing\relax M.~Dunion, T.~McInroe, K.~S. Luck, J.~P. Hanna, and S.~V. Albrecht, ``Temporal disentanglement of representations for improved generalisation in reinforcement learning,'' in \emph{The Eleventh International Conference on Learning Representations, {ICLR} 2023, Kigali, Rwanda, May 1-5, 2023}.\hskip 1em plus 0.5em minus 0.4em\relax OpenReview.net, 2023. [Online]. Available: \url{https://openreview.net/pdf?id=sPgP6aISLTD}\relax
\mciteBstWouldAddEndPunctfalse
\mciteSetBstMidEndSepPunct{~;\space}{}{\par\BIBentrySTDinterwordspacing}\relax
\EndOfBibitem
\bibitem{DBLP:conf/iclr/YaratsKF21}
\BIBentryALTinterwordspacing\relax D.~Yarats, I.~Kostrikov, and R.~Fergus, ``Image augmentation is all you need: Regularizing deep reinforcement learning from pixels,'' in \emph{9th International Conference on Learning Representations, {ICLR} 2021, Virtual Event, Austria, May 3-7, 2021}.\hskip 1em plus 0.5em minus 0.4em\relax OpenReview.net, 2021. [Online]. Available: \url{https://openreview.net/forum?id=GY6-6sTvGaf}\relax
\mciteBstWouldAddEndPunctfalse
\mciteSetBstMidEndSepPunct{~;\space}{}{\par\BIBentrySTDinterwordspacing}\relax
\EndOfBibitem
\bibitem{DBLP:conf/iclr/YaratsFLP22}
\BIBentryALTinterwordspacing\relax D.~Yarats, R.~Fergus, A.~Lazaric, and L.~Pinto, ``Mastering visual continuous control: Improved data-augmented reinforcement learning,'' in \emph{The Tenth International Conference on Learning Representations, {ICLR} 2022, Virtual Event, April 25-29, 2022}.\hskip 1em plus 0.5em minus 0.4em\relax OpenReview.net, 2022. [Online]. Available: \url{https://openreview.net/forum?id=\_SJ-\_yyes8}\relax
\mciteBstWouldAddEndPunctfalse
\mciteSetBstMidEndSepPunct{~;\space}{}{\par\BIBentrySTDinterwordspacing}\relax
\EndOfBibitem
\bibitem{DBLP:conf/nips/LaskinLSPAS20}
\BIBentryALTinterwordspacing\relax M.~Laskin, K.~Lee, A.~Stooke, L.~Pinto, P.~Abbeel, and A.~Srinivas, ``Reinforcement learning with augmented data,'' in \emph{Advances in Neural Information Processing Systems 33: Annual Conference on Neural Information Processing Systems 2020, NeurIPS 2020, December 6-12, 2020, virtual}, H.~Larochelle, M.~Ranzato, R.~Hadsell, M.~Balcan, and H.~Lin, Eds., 2020. [Online]. Available: \url{https://proceedings.neurips.cc/paper/2020/hash/e615c82aba461681ade82da2da38004a-Abstract.html}\relax
\mciteBstWouldAddEndPunctfalse
\mciteSetBstMidEndSepPunct{~;\space}{}{\par\BIBentrySTDinterwordspacing}\relax
\EndOfBibitem
\bibitem{DBLP:conf/aaai/00020CX022}
\BIBentryALTinterwordspacing\relax K.~Wu, M.~Wu, Z.~Chen, Y.~Xu, and X.~Li, ``Generalizing reinforcement learning through fusing self-supervised learning into intrinsic motivation,'' in \emph{Thirty-Sixth {AAAI} Conference on Artificial Intelligence, {AAAI} 2022, Thirty-Fourth Conference on Innovative Applications of Artificial Intelligence, {IAAI} 2022, The Twelveth Symposium on Educational Advances in Artificial Intelligence, {EAAI} 2022 Virtual Event, February 22 - March 1, 2022}.\hskip 1em plus 0.5em minus 0.4em\relax {AAAI} Press, 2022, pp. 8683--8690. [Online]. Available: \url{https://doi.org/10.1609/aaai.v36i8.20847}\relax
\mciteBstWouldAddEndPunctfalse
\mciteSetBstMidEndSepPunct{~;\space}{}{\par\BIBentrySTDinterwordspacing}\relax
\EndOfBibitem
\bibitem{DBLP:conf/cvpr/HeCXLDG22}
\BIBentryALTinterwordspacing\relax K.~He, X.~Chen, S.~Xie, Y.~Li, P.~Doll{\'{a}}r, and R.~B. Girshick, ``Masked autoencoders are scalable vision learners,'' in \emph{{IEEE/CVF} Conference on Computer Vision and Pattern Recognition, {CVPR} 2022, New Orleans, LA, USA, June 18-24, 2022}.\hskip 1em plus 0.5em minus 0.4em\relax {IEEE}, 2022, pp. 15\,979--15\,988. [Online]. Available: \url{https://doi.org/10.1109/CVPR52688.2022.01553}\relax
\mciteBstWouldAddEndPunctfalse
\mciteSetBstMidEndSepPunct{~;\space}{}{\par\BIBentrySTDinterwordspacing}\relax
\EndOfBibitem
\bibitem{DBLP:journals/corr/abs-2203-06173}
\BIBentryALTinterwordspacing\relax T.~Xiao, I.~Radosavovic, T.~Darrell, and J.~Malik, ``Masked visual pre-training for motor control,'' \emph{CoRR}, vol. abs/2203.06173, 2022. [Online]. Available: \url{https://doi.org/10.48550/arXiv.2203.06173}\relax
\mciteBstWouldAddEndPunctfalse
\mciteSetBstMidEndSepPunct{~;\space}{}{\par\BIBentrySTDinterwordspacing}\relax
\EndOfBibitem
\bibitem{DBLP:conf/nips/YuZLLC22}
\BIBentryALTinterwordspacing\relax T.~Yu, Z.~Zhang, C.~Lan, Y.~Lu, and Z.~Chen, ``Mask-based latent reconstruction for reinforcement learning,'' in \emph{Advances in Neural Information Processing Systems 35: Annual Conference on Neural Information Processing Systems 2022, NeurIPS 2022, New Orleans, LA, USA, November 28 - December 9, 2022}, S.~Koyejo, S.~Mohamed, A.~Agarwal, D.~Belgrave, K.~Cho, and A.~Oh, Eds., 2022. [Online]. Available: \url{http://papers.nips.cc/paper\_files/paper/2022/hash/a0709efe5139939ab69902884ecad9c1-Abstract-Conference.html}\relax
\mciteBstWouldAddEndPunctfalse
\mciteSetBstMidEndSepPunct{~;\space}{}{\par\BIBentrySTDinterwordspacing}\relax
\EndOfBibitem
\bibitem{DBLP:conf/iros/KotbWW23}
\BIBentryALTinterwordspacing\relax M.~Kotb, C.~Weber, and S.~Wermter, ``Sample-efficient real-time planning with curiosity cross-entropy method and contrastive learning,'' in \emph{{IROS}}, 2023, pp. 9456--9463. [Online]. Available: \url{https://doi.org/10.1109/IROS55552.2023.10342018}\relax
\mciteBstWouldAddEndPunctfalse
\mciteSetBstMidEndSepPunct{~;\space}{}{\par\BIBentrySTDinterwordspacing}\relax
\EndOfBibitem
\bibitem{DBLP:conf/iccv/WangWWTL21}
\BIBentryALTinterwordspacing\relax G.~Wang, K.~Wang, G.~Wang, P.~H.~S. Torr, and L.~Lin, ``Solving inefficiency of self-supervised representation learning,'' in \emph{2021 {IEEE/CVF} International Conference on Computer Vision, {ICCV} 2021, Montreal, QC, Canada, October 10-17, 2021}.\hskip 1em plus 0.5em minus 0.4em\relax {IEEE}, 2021, pp. 9485--9495. [Online]. Available: \url{https://doi.org/10.1109/ICCV48922.2021.00937}\relax
\mciteBstWouldAddEndPunctfalse
\mciteSetBstMidEndSepPunct{~;\space}{}{\par\BIBentrySTDinterwordspacing}\relax
\EndOfBibitem
\bibitem{DBLP:conf/iclr/ChenHTC022}
\BIBentryALTinterwordspacing\relax T.~Chen, W.~Hung, H.~Tseng, S.~Chien, and M.~Yang, ``Incremental false negative detection for contrastive learning,'' in \emph{The Tenth International Conference on Learning Representations, {ICLR} 2022, Virtual Event, April 25-29, 2022}.\hskip 1em plus 0.5em minus 0.4em\relax OpenReview.net, 2022. [Online]. Available: \url{https://openreview.net/forum?id=dDjSKKA5TP1}\relax
\mciteBstWouldAddEndPunctfalse
\mciteSetBstMidEndSepPunct{~;\space}{}{\par\BIBentrySTDinterwordspacing}\relax
\EndOfBibitem
\bibitem{DBLP:conf/nips/ChuangRL0J20}
\BIBentryALTinterwordspacing\relax C.~Chuang, J.~Robinson, Y.~Lin, A.~Torralba, and S.~Jegelka, ``Debiased contrastive learning,'' in \emph{Advances in Neural Information Processing Systems 33: Annual Conference on Neural Information Processing Systems 2020, NeurIPS 2020, December 6-12, 2020, virtual}, H.~Larochelle, M.~Ranzato, R.~Hadsell, M.~Balcan, and H.~Lin, Eds., 2020. [Online]. Available: \url{https://proceedings.neurips.cc/paper/2020/hash/63c3ddcc7b23daa1e42dc41f9a44a873-Abstract.html}\relax
\mciteBstWouldAddEndPunctfalse
\mciteSetBstMidEndSepPunct{~;\space}{}{\par\BIBentrySTDinterwordspacing}\relax
\EndOfBibitem
\bibitem{DBLP:conf/nips/KalantidisSPWL20}
\BIBentryALTinterwordspacing\relax Y.~Kalantidis, M.~B. Sariyildiz, N.~Pion, P.~Weinzaepfel, and D.~Larlus, ``Hard negative mixing for contrastive learning,'' in \emph{Advances in Neural Information Processing Systems 33: Annual Conference on Neural Information Processing Systems 2020, NeurIPS 2020, December 6-12, 2020, virtual}, H.~Larochelle, M.~Ranzato, R.~Hadsell, M.~Balcan, and H.~Lin, Eds., 2020. [Online]. Available: \url{https://proceedings.neurips.cc/paper/2020/hash/f7cade80b7cc92b991cf4d2806d6bd78-Abstract.html}\relax
\mciteBstWouldAddEndPunctfalse
\mciteSetBstMidEndSepPunct{~;\space}{}{\par\BIBentrySTDinterwordspacing}\relax
\EndOfBibitem
\bibitem{DBLP:conf/iclr/RobinsonCSJ21}
\BIBentryALTinterwordspacing\relax J.~D. Robinson, C.~Chuang, S.~Sra, and S.~Jegelka, ``Contrastive learning with hard negative samples,'' in \emph{9th International Conference on Learning Representations, {ICLR} 2021, Virtual Event, Austria, May 3-7, 2021}.\hskip 1em plus 0.5em minus 0.4em\relax OpenReview.net, 2021. [Online]. Available: \url{https://openreview.net/forum?id=CR1XOQ0UTh-}\relax
\mciteBstWouldAddEndPunctfalse
\mciteSetBstMidEndSepPunct{~;\space}{}{\par\BIBentrySTDinterwordspacing}\relax
\EndOfBibitem
\bibitem{DBLP:journals/corr/abs-2401-00165}
\BIBentryALTinterwordspacing\relax S.~Wang, Y.~Zhang, and C.~Nguyen, ``Mitigating the impact of false negatives in dense retrieval with contrastive confidence regularization,'' \emph{CoRR}, vol. abs/2401.00165, 2024. [Online]. Available: \url{https://doi.org/10.48550/arXiv.2401.00165}\relax
\mciteBstWouldAddEndPunctfalse
\mciteSetBstMidEndSepPunct{~;\space}{}{\par\BIBentrySTDinterwordspacing}\relax
\EndOfBibitem
\bibitem{DBLP:journals/ml/JordanGJS99}
\BIBentryALTinterwordspacing\relax M.~I. Jordan, Z.~Ghahramani, T.~S. Jaakkola, and L.~K. Saul, ``An introduction to variational methods for graphical models,'' \emph{Mach. Learn.}, vol.~37, no.~2, pp. 183--233, 1999. [Online]. Available: \url{https://doi.org/10.1023/A:1007665907178}\relax
\mciteBstWouldAddEndPunctfalse
\mciteSetBstMidEndSepPunct{~;\space}{}{\par\BIBentrySTDinterwordspacing}\relax
\EndOfBibitem
\bibitem{DBLP:conf/nips/BromleyGLSS93}
\BIBentryALTinterwordspacing\relax J.~Bromley, I.~Guyon, Y.~LeCun, E.~S{\"{a}}ckinger, and R.~Shah, ``Signature verification using a siamese time delay neural network,'' in \emph{Advances in Neural Information Processing Systems 6, [7th {NIPS} Conference, Denver, Colorado, USA, 1993]}, J.~D. Cowan, G.~Tesauro, and J.~Alspector, Eds.\hskip 1em plus 0.5em minus 0.4em\relax Morgan Kaufmann, 1993, pp. 737--744. [Online]. Available: \url{http://papers.nips.cc/paper/769-signature-verification-using-a-siamese-time-delay-neural-network}\relax
\mciteBstWouldAddEndPunctfalse
\mciteSetBstMidEndSepPunct{~;\space}{}{\par\BIBentrySTDinterwordspacing}\relax
\EndOfBibitem
\bibitem{DBLP:conf/icml/FujimotoHM18}
\BIBentryALTinterwordspacing\relax S.~Fujimoto, H.~van Hoof, and D.~Meger, ``Addressing function approximation error in actor-critic methods,'' in \emph{Proceedings of the 35th International Conference on Machine Learning, {ICML} 2018, Stockholmsm{\"{a}}ssan, Stockholm, Sweden, July 10-15, 2018}, ser. Proceedings of Machine Learning Research, J.~G. Dy and A.~Krause, Eds., vol.~80.\hskip 1em plus 0.5em minus 0.4em\relax {PMLR}, 2018, pp. 1582--1591. [Online]. Available: \url{http://proceedings.mlr.press/v80/fujimoto18a.html}\relax
\mciteBstWouldAddEndPunctfalse
\mciteSetBstMidEndSepPunct{~;\space}{}{\par\BIBentrySTDinterwordspacing}\relax
\EndOfBibitem
\bibitem{DBLP:conf/icml/HaarnojaZAL18}
\BIBentryALTinterwordspacing\relax T.~Haarnoja, A.~Zhou, P.~Abbeel, and S.~Levine, ``Soft actor-critic: Off-policy maximum entropy deep reinforcement learning with a stochastic actor,'' in \emph{Proceedings of the 35th International Conference on Machine Learning, {ICML} 2018, Stockholmsm{\"{a}}ssan, Stockholm, Sweden, July 10-15, 2018}, ser. Proceedings of Machine Learning Research, J.~G. Dy and A.~Krause, Eds., vol.~80.\hskip 1em plus 0.5em minus 0.4em\relax {PMLR}, 2018, pp. 1856--1865. [Online]. Available: \url{http://proceedings.mlr.press/v80/haarnoja18b.html}\relax
\mciteBstWouldAddEndPunctfalse
\mciteSetBstMidEndSepPunct{~;\space}{}{\par\BIBentrySTDinterwordspacing}\relax
\EndOfBibitem
\bibitem{tassa2018deepmind}
Y.~Tassa, Y.~Doron, A.~Muldal, T.~Erez, Y.~Li, D.~d.~L. Casas, D.~Budden, A.~Abdolmaleki, J.~Merel, A.~Lefrancq \emph{et~al.}, ``Deepmind control suite,'' \emph{arXiv preprint arXiv:1801.00690}, 2018\relax
\mciteBstWouldAddEndPuncttrue
\mciteSetBstMidEndSepPunct{;\space}{.}{\par\relax}\relax
\EndOfBibitem
\bibitem{DBLP:journals/corr/abs-1811-06032}
\BIBentryALTinterwordspacing\relax A.~Zhang, Y.~Wu, and J.~Pineau, ``Natural environment benchmarks for reinforcement learning,'' \emph{CoRR}, vol. abs/1811.06032, 2018. [Online]. Available: \url{http://arxiv.org/abs/1811.06032}\relax
\mciteBstWouldAddEndPunctfalse
\mciteSetBstMidEndSepPunct{~;\space}{}{\par\BIBentrySTDinterwordspacing}\relax
\EndOfBibitem
\bibitem{DBLP:conf/iclr/GuXLLLMTTWYYXHC23}
\BIBentryALTinterwordspacing\relax J.~Gu, F.~Xiang, X.~Li, Z.~Ling, X.~Liu, T.~Mu, Y.~Tang, S.~Tao, X.~Wei, Y.~Yao, X.~Yuan, P.~Xie, Z.~Huang, R.~Chen, and H.~Su, ``Maniskill2: {A} unified benchmark for generalizable manipulation skills,'' in \emph{The Eleventh International Conference on Learning Representations, {ICLR} 2023, Kigali, Rwanda, May 1-5, 2023}.\hskip 1em plus 0.5em minus 0.4em\relax OpenReview.net, 2023. [Online]. Available: \url{https://openreview.net/forum?id=b\_CQDy9vrD1}\relax
\mciteBstWouldAddEndPunctfalse
\mciteSetBstMidEndSepPunct{~;\space}{}{\par\BIBentrySTDinterwordspacing}\relax
\EndOfBibitem
\bibitem{DBLP:conf/nips/ZhuSG023}
\BIBentryALTinterwordspacing\relax C.~Zhu, M.~Simchowitz, S.~Gadipudi, and A.~Gupta, ``Repo: Resilient model-based reinforcement learning by regularizing posterior predictability,'' in \emph{Advances in Neural Information Processing Systems 36: Annual Conference on Neural Information Processing Systems 2023, NeurIPS 2023, New Orleans, LA, USA, December 10 - 16, 2023}, A.~Oh, T.~Naumann, A.~Globerson, K.~Saenko, M.~Hardt, and S.~Levine, Eds., 2023. [Online]. Available: \url{http://papers.nips.cc/paper\_files/paper/2023/hash/6692e1b0e8a31e8de84bd90ad4d8d9e0-Abstract-Conference.html}\relax
\mciteBstWouldAddEndPunctfalse
\mciteSetBstMidEndSepPunct{~;\space}{}{\par\BIBentrySTDinterwordspacing}\relax
\EndOfBibitem
\bibitem{DBLP:conf/icml/SunZ0I24}
\BIBentryALTinterwordspacing\relax R.~Sun, H.~Zang, X.~Li, and R.~Islam, ``Learning latent dynamic robust representations for world models,'' in \emph{Forty-first International Conference on Machine Learning, {ICML} 2024, Vienna, Austria, July 21-27, 2024}.\hskip 1em plus 0.5em minus 0.4em\relax OpenReview.net, 2024. [Online]. Available: \url{https://openreview.net/forum?id=C4jkx6AgWc}\relax
\mciteBstWouldAddEndPunctfalse
\mciteSetBstMidEndSepPunct{~;\space}{}{\par\BIBentrySTDinterwordspacing}\relax
\EndOfBibitem
\bibitem{DBLP:conf/icml/PooleOOAT19}
\BIBentryALTinterwordspacing\relax B.~Poole, S.~Ozair, A.~van~den Oord, A.~A. Alemi, and G.~Tucker, ``On variational bounds of mutual information,'' in \emph{Proceedings of the 36th International Conference on Machine Learning, {ICML} 2019, 9-15 June 2019, Long Beach, California, {USA}}, ser. Proceedings of Machine Learning Research, K.~Chaudhuri and R.~Salakhutdinov, Eds., vol.~97.\hskip 1em plus 0.5em minus 0.4em\relax {PMLR}, 2019, pp. 5171--5180. [Online]. Available: \url{http://proceedings.mlr.press/v97/poole19a.html}\relax
\mciteBstWouldAddEndPunctfalse
\mciteSetBstMidEndSepPunct{~;\space}{}{\par\BIBentrySTDinterwordspacing}\relax
\EndOfBibitem
\bibitem{DBLP:journals/neco/LeCunBDHHHJ89}
\BIBentryALTinterwordspacing\relax Y.~LeCun, B.~E. Boser, J.~S. Denker, D.~Henderson, R.~E. Howard, W.~E. Hubbard, and L.~D. Jackel, ``Backpropagation applied to handwritten zip code recognition,'' \emph{Neural Comput.}, vol.~1, no.~4, pp. 541--551, 1989. [Online]. Available: \url{https://doi.org/10.1162/neco.1989.1.4.541}\relax
\mciteBstWouldAddEndPunctfalse
\mciteSetBstMidEndSepPunct{~;\space}{}{\par\BIBentrySTDinterwordspacing}\relax
\EndOfBibitem
\bibitem{DBLP:conf/emnlp/ChoMGBBSB14}
\BIBentryALTinterwordspacing\relax K.~Cho, B.~van Merrienboer, {\c{C}}.~G{\"{u}}l{\c{c}}ehre, D.~Bahdanau, F.~Bougares, H.~Schwenk, and Y.~Bengio, ``Learning phrase representations using {RNN} encoder-decoder for statistical machine translation,'' in \emph{Proceedings of the 2014 Conference on Empirical Methods in Natural Language Processing, {EMNLP} 2014, October 25-29, 2014, Doha, Qatar, {A} meeting of SIGDAT, a Special Interest Group of the {ACL}}, A.~Moschitti, B.~Pang, and W.~Daelemans, Eds.\hskip 1em plus 0.5em minus 0.4em\relax {ACL}, 2014, pp. 1724--1734. [Online]. Available: \url{https://doi.org/10.3115/v1/d14-1179}\relax
\mciteBstWouldAddEndPunctfalse
\mciteSetBstMidEndSepPunct{~;\space}{}{\par\BIBentrySTDinterwordspacing}\relax
\EndOfBibitem
\bibitem{DBLP:journals/corr/KingmaB14}
\BIBentryALTinterwordspacing\relax D.~P. Kingma and J.~Ba, ``Adam: {A} method for stochastic optimization,'' in \emph{3rd International Conference on Learning Representations, {ICLR} 2015, San Diego, CA, USA, May 7-9, 2015, Conference Track Proceedings}, Y.~Bengio and Y.~LeCun, Eds., 2015. [Online]. Available: \url{http://arxiv.org/abs/1412.6980}\relax
\mciteBstWouldAddEndPunctfalse
\mciteSetBstMidEndSepPunct{~;\space}{}{\par\BIBentrySTDinterwordspacing}\relax
\EndOfBibitem
\end{mcitethebibliography}
\bibliographystyle{IEEEtranM}

\newpage

 

% \end{IEEEbiography}

% \begin{IEEEbiography}[{\includegraphics[width=1in,height=1.25in,clip,keepaspectratio]{figures/Shiguang Sun.JPG}}]{Shiguang Sun}

% \end{IEEEbiography}


% \vfill

\end{document}